\documentclass[twocolumn]{article}

\usepackage{arxiv}

\usepackage[utf8]{inputenc} 
\usepackage[T1]{fontenc}   

\usepackage{titletoc}

\usepackage{hyperref}   
\hypersetup{hidelinks}

\usepackage{xurl}
\usepackage{booktabs}       
\usepackage{amsfonts}       
\usepackage{nicefrac}       
\usepackage{microtype}      
\usepackage{lipsum}		
\usepackage{graphicx}

\usepackage{doi}

\usepackage{mathtools}

\usepackage{amsmath,amssymb}
\usepackage{amsthm}

\theoremstyle{plain}
\newtheorem{theorem}{Theorem}
\newtheorem{lemma}{Lemma}
\newtheorem{proposition}{Proposition}

\theoremstyle{remark}
\newtheorem{remark}{Remark}

\usepackage[numbers,sort&compress]{natbib}
\usepackage[capitalize,noabbrev]{cleveref}
\usepackage{subcaption}

\title{Geometric Decomposition of Generative Feature Shift with an Application to Classifier-Free Guidance Scheduling}

\date{} 				

\author{
    \href{https://orcid.org/0009-0007-1946-9481}{%
        \includegraphics[scale=0.06]{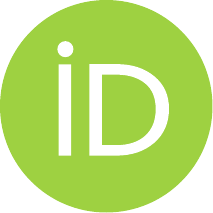}\hspace{1mm}Jihyoung Yi}
    \qquad
    \href{https://orcid.org/0000-0003-4413-4441}{%
        \includegraphics[scale=0.06]{orcid.pdf}\hspace{1mm}Zhou Wang}\\
    Department of Electrical and Computer Engineering\\
    University of Waterloo\\
    Waterloo, ON N2L 3G1, Canada\\
    \texttt{jihyoung.yi@uwaterloo.ca}
    \quad
    \texttt{zhou.wang@uwaterloo.ca}
}

\renewcommand{\shorttitle}{Geometric Decomposition of Generative Feature Shift}

\hypersetup{
  pdftitle={Geometric Decomposition of Generative Feature Shift with an Application to Classifier-Free Guidance Scheduling},  
  pdfauthor={Jihyoung Yi and Zhou Wang},
  pdfsubject={Generative feature shift and Bures--Wasserstein geometry},
  pdfkeywords={Bures--Wasserstein geometry, generative models, classifier-free guidance}
}

\begin{document}

\twocolumn[{%
  \maketitle
  \begin{abstract}
Generative interventions reshape feature distributions in ways that distance alone does not capture: two image collections can lie equally far from a reference yet redistribute feature variation differently. We decompose the squared Bures--Wasserstein (BW) distance between feature second-moment operators into spectral and organizational energies. These measure displacement within and across reference eigenspaces, respectively, and sum exactly to the squared distance. For trace-one operators, we derive a lower bound on spectral energy determined by distance alone. In controlled Gaussian mixtures at comparable BW distances, coverage imbalance produces predominantly spectral displacement, while mode merging and spurious-mode addition produce predominantly organizational displacement at the tested bandwidth. On CIFAR-10, we predict whether incoming images belong to a class already present in the reference collection or to a previously absent class. At fixed replacement fractions, adding spectral allocation features to a DINOv2 baseline with BW distance, Vendi diversity, Gaussian MMD$^2$, and replacement-weighted CMMD reduces held-out Brier error by $16.6\%$, with only a small predictive gain at matched BW distance. Spectral share also falls as the reference contains more distinct images per class, under both fixed replacement and matched BW distance. In SDXL, tracking spectral energy above the lower bound identifies an early response to classifier-free guidance and suggests a schedule to test. With only five guided updates in each schedule, applying guidance at the start rather than the end of the middle-noise interval increases final requested-category detection from $43.9\%$ to $68.2\%$. Code is available at \url{https://github.com/jihyoungyi/bw-geometry}.
\end{abstract}
  \vspace{1em}
}]

\section{Introduction}
\label{sec:intro}

Generative interventions such as classifier-free guidance (CFG) and truncation in generative adversarial networks (GANs) reshape generated distributions \cite{ho_classifier-free_2021, karras_style-based_2021}. Common generative evaluation metrics, such as the Fr\'{e}chet Inception Distance (FID) \cite{heusel_gans_2017} and precision, recall, density, and coverage (PRDC) \cite{kynkaanniemi_improved_2019, naeem_reliable_2020}, summarize distributional discrepancy, fidelity, and diversity. However, these scores do not provide an exact additive decomposition of the observed distributional shift into distinct geometric contributions. Analytical studies provide mechanistic insight into how generative interventions reshape distributions under explicit modeling assumptions \cite{humayun_polarity_2022, chidambaram_what_2024, li_towards_2025, jin_stage-wise_2026}. This motivates a complementary, sample-based diagnostic for characterizing shifts in feature distributions. We introduce such a diagnostic by decomposing feature second-moment displacement into interpretable geometric components. 

Our decomposition yields \emph{spectral} and \emph{organizational} displacement components that are orthogonal under the Bures--Wasserstein (BW) metric \cite{bhatia_bureswasserstein_2019} at the reference operator. The spectral component lies within the reference eigenspaces, while the organizational component lies across distinct reference eigenspaces. Our construction combines the classical Bures-orthogonal splitting of tangent space into reference-commuting and orbit directions \cite{dittmann_explicit_1999} with the BW geodesic geometry of positive-definite matrices \cite{bhatia_bureswasserstein_2019}. We apply this splitting to the exact BW logarithmic displacement between arbitrary positive-definite endpoints, yielding an additive energy allocation without an infinitesimal-displacement approximation. For feature covariance matrices, the squared BW distance equals FID's covariance term \cite{dowson_frechet_1982, bhatia_bureswasserstein_2019}. We use normalized, regularized uncentered second-moment operators without assuming Gaussian feature distributions; see \cref{app:fid_relation} for details. For trace-one operators, we show that the squared BW distance determines a lower bound on spectral energy (\cref{thm:spectral-trace-one}). At matched distance, spectral-share differences therefore reflect allocation above a common bound. When distance changes during denoising, tracking spectral energy above the bound tells us how much within-eigenspace displacement remains after accounting for the contribution required by distance alone.

To demonstrate the diagnostic's empirical utility, we investigate two questions. First, can the allocation help distinguish the introduction of a previously absent class from new examples of a familiar class? We test this on CIFAR-10, holding either the replacement fraction or the BW distance fixed (\cref{sec:controlled_experiments}). Second, can tracking when CFG changes the allocation of displacement inform when guidance should be applied during denoising? Previous studies investigate how guidance strength and timing affect conditional alignment, sample diversity, and sampling dynamics \cite{kynkaanniemi_applying_2024, li_towards_2025, jin_stage-wise_2026}. In Stable Diffusion XL (SDXL) \cite{podell_sdxl_2024}, we use the diagnostic to motivate a comparison of early and late guidance within the same middle-noise interval (\cref{sec:stable_diffusion_exp}). 

Our contributions are twofold: 

\textbf{(i) An exact diagnostic of finite feature shift:} For a specified reference, the two BW-orthogonal projections give a unique allocation of energy. The component energies sum exactly to the squared BW distance for arbitrary positive-definite endpoints and are invariant under common orthogonal changes of feature coordinates. For trace-one operators, we also derive a distance-dependent lower bound on spectral energy. 

\begin{figure}[t]
    \centering
    \begin{subfigure}[t]{\linewidth}
        \centering
        \includegraphics[width=\linewidth]
            {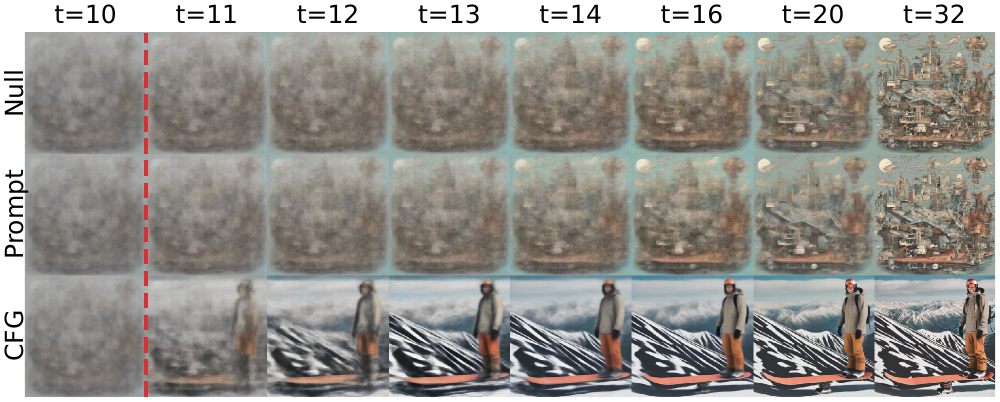}
        \caption{Denoising trajectories for
        ``a photo of a person and a snowboard.''}
        \label{fig:cfg_switch_example}
    \end{subfigure}\par\smallskip
    \begin{subfigure}[t]{\linewidth}
        \centering
        \includegraphics[width=\linewidth]
            {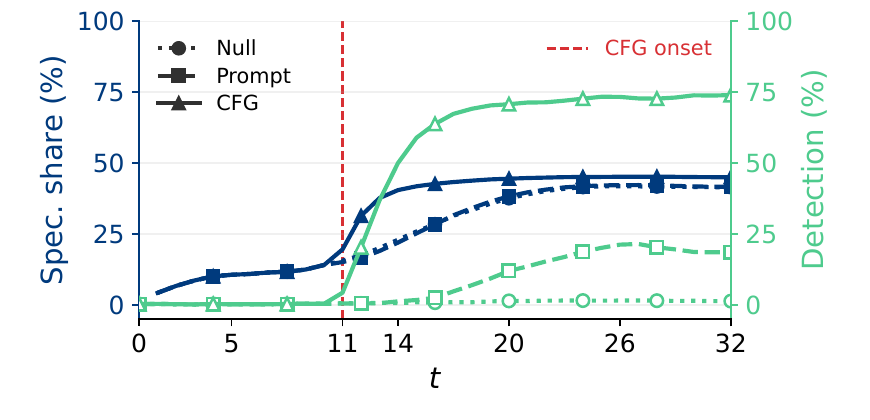}
        \caption{Spectral share and requested-category detection.}
        \label{fig:cfg_switch_share_detection}
    \end{subfigure}
    \caption{\textbf{Feature shift at guidance onset in SDXL.}
    (a) The three branches share the same latent through $t=10$. The red dashed line marks the first guided update.
    (b) Filled blue markers show spectral share relative to the fixed state-0 reference (left axis); hollow green markers show the fraction of requested categories detected (right axis). Marker shapes and line styles identify the
    branches. Curves summarize 40 prompts with 32 seeds each.}
    \label{fig:cfg_switch_timeline}
\end{figure}

\textbf{(ii) Validating and interpreting the allocation across distribution shifts and denoising:} Controlled Gaussian-mixture experiments demonstrate that shifts with comparable BW distances can exhibit sharply different spectral and organizational allocations under a fixed feature map. On CIFAR-10 \cite{krizhevsky_learning_2009}, introducing a previously absent class yields higher mean spectral share than adding examples of a familiar class at matched BW distance in DINOv2 \cite{oquab_dinov2_2024} features. Spectral share decreases as reference collections contain more distinct images per class. It also adds predictive information beyond distance and diversity baselines, with the clearest benefit under fixed replacement. In SDXL \cite{podell_sdxl_2024}, the allocation responds to CFG before widespread requested-category detection (\cref{fig:cfg_switch_timeline}). Tracking spectral energy above its distance-dependent lower bound motivates an equal-budget scheduling test: applying guidance during the first five rather than the last five updates of the middle-noise interval improves final requested-category detection by $24.3$ percentage points.

\section{Related Work}
\label{sec:related_work}

Generative-model evaluation commonly compares samples in pretrained feature spaces \cite{stein_exposing_2023}. FID \cite{heusel_gans_2017} compares feature means and covariances under a Gaussian approximation, while CMMD \cite{jayasumana_rethinking_2024} uses maximum mean discrepancy with a Gaussian kernel on CLIP features \cite{radford_learning_2021}. In our CIFAR-10 comparisons, we compute Gaussian-kernel MMD$^2$ on the same features used for the BW decomposition. PRDC \cite{kynkaanniemi_improved_2019,naeem_reliable_2020} characterize fidelity and support coverage through local neighborhoods. These metrics provide complementary assessments rather than an additive decomposition of a common discrepancy. Our diagnostic instead allocates a single total discrepancy---the squared BW distance---into reference-relative spectral and organizational energies that sum exactly to that total. R\'{e}nyi Kernel Entropy (RKE) \cite{jalali_information-theoretic_2023} and the Vendi Score \cite{friedman_vendi_2023} quantify diversity through kernel or similarity spectra. \citet{jalali_information-theoretic_2025} decompose the Vendi Score into the product of Conditional-Vendi and Information-Vendi scores, distinguishing diversity remaining after conditioning on prompts from diversity associated with differences between prompts. Relative R\'{e}nyi Kernel Entropy (RRKE) \cite{jalali_information-theoretic_2023} is a particularly close antecedent: it compares trace-one uncentered second-moment operators in kernel feature space and, at order $\frac{1}{2}$, is a strictly increasing transform of their squared BW distance. In our real-image experiments, we instantiate this operator construction using normalized pretrained features and compare reference and intervention conditions. Our contribution is the additive allocation of squared BW distance into reference-relative spectral and organizational energies. 

Classical Bures Riemannian geometry gives a BW-orthogonal tangent-space splitting into directions commuting with the reference matrix and directions tangent to its isospectral orbit \cite{dittmann_explicit_1999}; finite BW distances and geodesics are also established \cite{bhatia_bureswasserstein_2019}. We project the exact BW logarithmic displacement onto these two subspaces, separating changes within each reference eigenspace from changes across distinct reference eigenspaces. Each projection is the unique best approximation within its respective subspace in the BW tangent norm. Their squared norms define spectral and organizational energies that sum exactly to squared BW distance for arbitrary positive-definite endpoint pairs, without an infinitesimal-displacement approximation. BW geometry also appears in neural-representation comparison: \citet{harvey_duality_2024} connect normalized Bures similarity between centered linear stimulus kernel matrices to shape distances between representations. Their setting compares neural representations of shared inputs across networks; ours compares distributions in a fixed feature space and decomposes the resulting operator displacement. Commutator norms provide related measures of noncommutativity and connect to approximate joint diagonalization \cite{glashoff_matrix_2013}.

CFG typically trades stronger conditional adherence for reduced diversity \cite{ho_classifier-free_2021}, with effects that depend on timing. \citet{sadat_cads_2024} improve sample diversity by perturbing the conditioning signal with Gaussian noise that is annealed during sampling. \citet{kynkaanniemi_applying_2024} show that restricting guidance to an intermediate noise interval can improve the fidelity--diversity trade-off. We examine when CFG changes the allocation of feature displacement within that interval. Theoretical work characterizes both guided dynamics and induced distributions. In mixture-model settings, \citet{chidambaram_what_2024} show that strong guidance drives samples toward extreme regions of the conditional distribution. Under their Gaussian-mixture assumptions, \citet{jin_stage-wise_2026} characterize high-noise direction shift, intermediate mode separation, and low-noise within-mode concentration, including how early guidance bias suppresses access to weaker modes. Among analyses of CFG, the decomposition of \citet{li_towards_2025} is most closely related to ours. They decompose the CFG guidance term in linear diffusion models into a mean-shift term and positive and negative contrastive principal component (CPC) terms, and investigate their effects in nonlinear models. The CPC terms amplify or suppress directions in the positive or negative eigenspaces, respectively, of a noise-dependent contrast derived from conditional and unconditional covariances. Whereas their decomposition describes the CFG guidance, we decompose the resulting feature displacement relative to a fixed reference during denoising.

\section{A Geometric Diagnostic for Generative Feature Shift}
\label{sec:geometric_diagnostic}

We seek to characterize the geometry of feature shift relative to a chosen baseline. Our diagnostic assigns the squared BW distance between two feature operators to displacements within and across the reference eigenspaces. We first specify the operators, then construct the finite displacement and its energy allocation. The construction applies independently of the generator or intervention; CFG provides the principal scientific application in \cref{sec:stable_diffusion_exp}. Proofs of the results are given in \cref{{app:proofs}} of the appendix.

\subsection{Representing distributions by feature operators}
\label{subsec:feature_distribution_operators}

Let \(\varphi:\mathcal X\to\mathbb R^d\) be a fixed, pointwise \(\ell_2\)-normalized feature map shared by the distributions being compared. For samples \(x_1,\ldots,x_n\sim P\), define
\begin{equation}
\label{eq:empirical-density-operator}
\widehat M_P=\frac1n\sum_{i=1}^n\varphi(x_i)\varphi(x_i)^\top,
\qquad
\widehat\rho(P)=\frac{\widehat M_P}{\operatorname{tr}(\widehat M_P)}.
\end{equation}
Pointwise normalization prevents feature magnitude from reweighting observations and gives $\operatorname{tr}(\widehat M_P)=1$ in exact arithmetic. We retain explicit trace normalization to enforce this constraint numerically. This uncentered second-moment operator retains both covariance and mean-direction structure, since at the population level \(M_P=\Sigma_P+\mu_P\mu_P^\top\). It summarizes second-order structure in the chosen feature space and does not identify the full underlying distribution. No Gaussian assumption is required. 

To ensure positive definiteness for the BW Riemannian constructions \cite{bhatia_bureswasserstein_2019}, we apply trace-preserving isotropic shrinkage,
\begin{equation}
\label{eq:diagnostic-shrinkage}
\widehat\rho_\delta(P)=(1-\delta)\widehat\rho(P)+\delta I/d,
\qquad 0<\delta<1.
\end{equation}
For fixed $d$ and $\delta$, shrinkage keeps the operators uniformly away from singularity and bounds the BW tangent norm
(\cref{app:regularized_operator_bounds}). This is a trace-normalized form of scaled-identity shrinkage \citep{ledoit_well-conditioned_2004,raninen_coupled_2021}. Here, shrinkage is used to obtain positive-definite, numerically admissible operators for BW geometry, rather than to minimize covariance-estimation error. We select the smallest positive grid candidate \(\delta_{\min}\) satisfying a numerical-conditioning criterion on an independent calibration set. Across all experiments, we use \(\delta_{\mathrm{used}}=10\times\delta_{\min}\) as a numerical safety margin. Both values ensure positive definiteness; the larger value improves conditioning by increasing the smallest eigenvalues, while assigning more weight to the isotropic target. Calibration is performed separately for each experimental setting and feature map, and the resulting value is held fixed across comparisons. \cref{app:shrinkage_selection,app:calibration_sets,app:eigenvalue_grouping} provide details of the selection rule, calibration sets, and numerical checks. Below, \(A=\widehat\rho_\delta(P_{\mathrm{ref}})\) denotes the reference and \(B=\widehat\rho_\delta(P_{\mathrm{int}})\) the intervention operator.

\subsection{An exact displacement between finite endpoints}
\label{sec:bures-distance-construction}

We summarize the BW constructions needed below \citep{bhatia_bureswasserstein_2019, van_oostrum_bureswasserstein_2022}. Let \(A\) and \(B\) lie on \(\mathbb S_{++}^{d}\), the symmetric positive-definite manifold. The squared BW distance is

\begin{equation}
\label{eq:intro-bures-distance}
\begin{aligned}
d_{\mathrm{BW}}^2(A,B)
&= \operatorname{tr}(A)+\operatorname{tr}(B) \\
&\quad -2\operatorname{tr}\!\left[
    (A^{1/2}BA^{1/2})^{1/2}
\right].
\end{aligned}
\end{equation}

The BW transport map from $A$ to $B$ is given by 
\begin{equation}
T_{A\to B}
=
A^{-1/2}
\left(A^{1/2}BA^{1/2}\right)^{1/2}
A^{-1/2}.
\end{equation} Differentiating the minimizing BW geodesic at $t=0$ yields its Riemannian logarithmic displacement:
\begin{equation}
\label{eq:intro-bures-log}
\begin{aligned}
V_{A\to B}
&:= \operatorname{Log}^{\mathrm{BW}}_A(B) \\
&= A(T_{A\to B}-I)+(T_{A\to B}-I)A.
\end{aligned}
\end{equation}

For tangent vectors \(H,K\in T_A\mathbb S_{++}^d\), set \(\mathcal L_A(X)=AX+XA\). The BW Riemannian metric and its induced norm are
\begin{equation}
\label{eq:diagnostic-bw-metric}
\begin{aligned}
g_A(H,K)
&= \tfrac12\operatorname{tr}\!\left[
    H\mathcal L_A^{-1}(K)
\right], \\
\|H\|_A^2
&= g_A(H,H).
\end{aligned}
\end{equation}

For arbitrary positive-definite endpoints, the exact identity
\begin{equation}
\label{eq:intro-log-norm-distance}
\|V_{A\to B}\|_A^2=d_{\mathrm{BW}}^2(A,B)
\end{equation}
ensures that decomposing \(V_{A\to B}\) orthogonally under the BW metric allocates the full squared distance without treating \(B-A\) as an infinitesimal perturbation. We use these established BW constructions \citep{bhatia_bureswasserstein_2019,van_oostrum_bureswasserstein_2022} in the ambient positive-definite geometry, even for trace-one endpoints, whereas the intrinsic Riemannian distance on the trace-one submanifold is the BW angle \citep{van_oostrum_bureswasserstein_2022}.

\subsection{Spectral and organizational energies}
\label{sec:intro-spectral-organizational}

We obtain a finite-pair decomposition by applying the classical BW-orthogonal commuting--orbit projections \citep{dittmann_explicit_1999} to the exact BW logarithmic displacement \(V_{A\to B}\) \citep{bhatia_bureswasserstein_2019}. These projections separate tangent displacement within the reference eigenspaces from displacement across distinct reference eigenspaces. 

The spectral projection is the pinching map with respect to the reference operator $A$ \cite{sutter_approximate_2018}; the organizational projection is its complement. Let \(A=\sum_{r=1}^{m}\alpha_rP_r\) be its spectral decomposition, where \(\alpha_1,\ldots,\alpha_m\) are the distinct eigenvalues of \(A\) and \(P_r\) projects onto the eigenspace associated with \(\alpha_r\). Using full eigenspaces makes the construction invariant to the choice of basis within any repeated eigenspace. In practice, we group adjacent reference eigenvalues whose difference falls within a numerical tolerance. For \(H\in T_A\mathbb S_{++}^d\), define the reference-anchored projections

\begin{equation}
\label{eq:diagnostic-projections}
\begin{aligned}
\Pi_A^{\mathrm{spec}}(H)
&\coloneqq \sum_r P_r H P_r, \\
\Pi_A^{\mathrm{org}}(H)
&\coloneqq H-\Pi_A^{\mathrm{spec}}(H) \\
&= \sum_{r\ne s} P_r H P_s.
\end{aligned}
\end{equation}

The spectral projection \(\Pi_A^{\mathrm{spec}}(H)\) retains within-eigenspace blocks, and the organizational projection \(\Pi_A^{\mathrm{org}}(H)\) retains cross-eigenspace blocks. Their ranges, \(\mathcal C_A:=\operatorname{range}(\Pi_A^{\mathrm{spec}})\) and \(\mathcal O_A:=\operatorname{range}(\Pi_A^{\mathrm{org}})\), are the reference-commuting and cross-eigenspace tangent subspaces, respectively. Applying these projections to \(V=V_{A\to B}\), we define the displacement components and their squared BW norms:

\begin{equation}
\label{eq:diagnostic-component-energies}
\begin{aligned}
V_{\mathrm{spec}}
&\coloneqq \Pi_A^{\mathrm{spec}}(V), \quad E_{\mathrm{spec}\mid A}(B)\coloneqq \|V_{\mathrm{spec}}\|_A^2, \\
V_{\mathrm{org}}
&\coloneqq \Pi_A^{\mathrm{org}}(V), \quad
E_{\mathrm{org}\mid A}(B)\coloneqq \|V_{\mathrm{org}}\|_A^2.
\end{aligned}
\end{equation}

\begin{theorem}[Exact allocation and best approximation]
\label{thm:finite-pair-decomposition}
For every \(A,B\in\mathbb S_{++}^d\), \(V=V_{\mathrm{spec}}+V_{\mathrm{org}}\) is a \(g_A\)-orthogonal decomposition, with nonnegative energies satisfying
\begin{equation}
\label{eq:finite-pair-pythagoras}
d_{\mathrm{BW}}^2(A,B)
= E_{\mathrm{spec}\mid A}(B)
+ E_{\mathrm{org}\mid A}(B).
\end{equation}
Moreover,
\begin{equation}
\label{eq:diagnostic-best-approximation}
\begin{aligned}
E_{\mathrm{org}\mid A}(B)
&= \min_{H\in\mathcal C_A}\|V-H\|_A^2, \\
E_{\mathrm{spec}\mid A}(B)
&= \min_{G\in\mathcal O_A}\|V-G\|_A^2,
\end{aligned}
\end{equation}
with unique minimizers \(H=V_{\mathrm{spec}}\) and \(G=V_{\mathrm{org}}\), respectively.
\end{theorem}
Thus, each component is the closest approximation to the full logarithmic displacement in its respective subspace under the BW norm.

\begin{proposition}[Commuting zero condition and coordinate invariance]
\label{prop:organizational-faithfulness}
For every \(A,B\in\mathbb S_{++}^d\),
\begin{equation}
\label{eq:diagnostic-commuting-zero}
E_{\mathrm{org}\mid A}(B)=0\quad\Longleftrightarrow\quad[A,B]=0.
\end{equation}
For any orthogonal \(U\), setting \(A'=UAU^\top\) and \(B'=UBU^\top\) gives
\begin{equation}
\label{eq:diagnostic-invariance}
E_{k\mid A'}(B')=E_{k\mid A}(B),\qquad k\in\{\mathrm{spec},\mathrm{org}\}.
\end{equation}
\end{proposition}
\begin{remark}
Thus, zero organizational energy means that \(B\) preserves each complete reference eigenspace, and a common orthogonal change of feature coordinates does not alter the allocation. However, this coordinate invariance does not imply independence from the chosen feature map: changing the encoder can change the diagnostic. The construction also recovers the classical commuting--orbit energies in the infinitesimal limit. 
\end{remark}

\begin{theorem}[Spectral energy for trace-one operators]
\label{thm:spectral-trace-one}
Let $A,B\in\mathbb{S}_{++}^{d}$ satisfy $\operatorname{tr}(A)=\operatorname{tr}(B)=1$, and write $D=d_{\mathrm{BW}}^2(A,B)$ for their squared ambient BW distance. Let $V=\operatorname{Log}^{\mathrm{BW}}_A(B)$ and $V_{\mathrm{spec}}=\Pi_A^{\mathrm{spec}}(V)$. Then
\begin{equation}
    E_{\mathrm{spec}\mid A}(B)
    = \frac{D^2}{4}
      + \bigl\|V_{\mathrm{spec}}+DA\bigr\|_A^2,
    \label{eq:spectral-trace-one}
\end{equation}
where $H:=V_{\mathrm{spec}}+DA$ has zero trace and commutes with $A$. Consequently, if $A\neq B$, the spectral share satisfies
\begin{equation}
    \eta_{\mathrm{spec}\mid A}(B)
    := \frac{E_{\mathrm{spec}\mid A}(B)}{D}
    \geq \frac{D}{4} > 0.
    \label{eq:spectral-share-floor}
\end{equation}
Equality holds if and only if $H=0$, equivalently
$V_{\mathrm{spec}}=-DA$.
\end{theorem}
The distinction between share and energy matters in our two experiments. In CIFAR-10, matching $D$ holds the share floor $D/4$ fixed, so a gap between spectral shares reflects a difference in the squared BW norm of trace-free displacement within the reference eigenspaces. In our SDXL CFG experiment, $D$ varies along the denoising trajectory. A rise in spectral share alone therefore does not establish an increase in spectral energy above its distance-dependent floor. We track this excess energy as $E_{\mathrm{spec}\mid A}(B)-D^2/4$.

\begin{remark}[Reference dependence and finite-displacement interpretation]
Although the total BW distance is symmetric, its component energies can change when $A$ and $B$ are exchanged: both the logarithm and the projections use the chosen reference. The energies quantify displacement within and across reference eigenspaces, rather than separate changes in endpoint eigenvalues and eigenvectors. For example, let $B=UAU^\top\ne A$ for an orthogonal $U$, with $\operatorname{tr}(A)=1$. The endpoints have identical eigenvalues, yet $D=d_{\mathrm{BW}}^2(A,B)>0$, so \cref{thm:spectral-trace-one} gives $E_{\mathrm{spec}\mid A}(B)\ge D^2/4>0$. Thus, spectral energy need not vanish when the endpoint spectra coincide. The two subspaces can differ in dimension, and the energies are not dimension-normalized; equal energies, therefore, do not constitute a default statistical baseline.
\end{remark}
\section{Controlled Validation}
\label{sec:controlled_experiments}

\paragraph{Gaussian mixtures:} We first test whether controlled shifts with approximately matched BW distances can produce different energy allocations. Motivated by Gaussian-mixture evaluations of generative diversity \citep{jalali_information-theoretic_2023}, we construct an eight-mode reference in two dimensions and apply three interventions under a shared, pointwise-normalized RBF feature map; \cref{app:gaussian_mixture} of the appendix provides additional details. The squared BW distances range from $0.074$ to $0.080$, yet spectral shares range from $3.74\%$ to $99.10\%$ (\cref{tab:synthetic_allocation}). At this feature-map setting, coverage imbalance is almost entirely spectral, whereas mode merging and spurious-mode addition are predominantly organizational.

\begin{table}[t]
    \centering
    \caption{\textbf{Energy allocation for controlled 2D Gaussian-mixture
    shifts.} Spectral share is the percentage of squared BW distance
    allocated within reference eigenspaces.}
    \label{tab:synthetic_allocation}
    \small
    \begin{tabular}{@{}lrr@{}}
        \hline
        Intervention & $d_{\mathrm{BW}}^2$ & Spectral share (\%) \\
        \hline
        Coverage imbalance     & 0.07602 & 99.10 \\
        Mode merging           & 0.07982 &  6.79 \\
        Spurious-mode addition & 0.07432 &  3.74 \\
        \hline
    \end{tabular}
\end{table}

\begin{figure}[t]
    \centering
    \begin{subfigure}[t]{\linewidth}
        \centering
        \includegraphics[width=\linewidth]
            {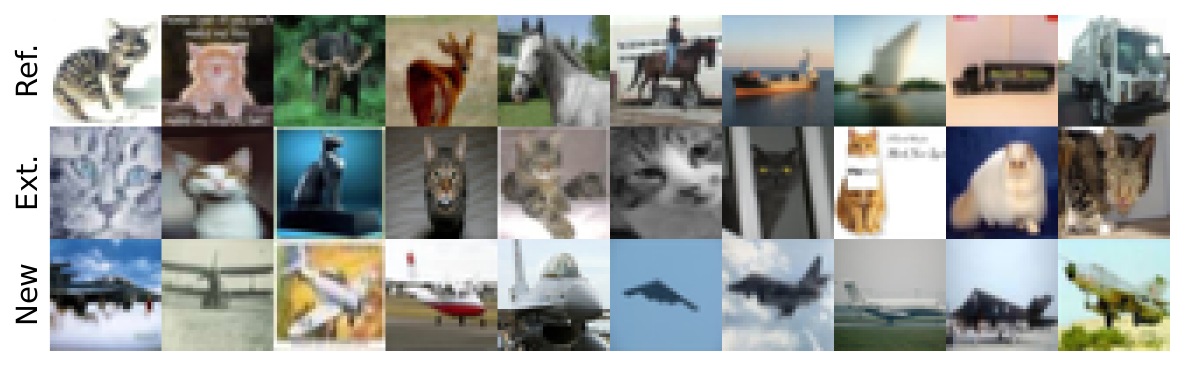}
        \caption{Reference and incoming images.}
        \label{fig:cifar_replacement_samples}
    \end{subfigure}\par\smallskip

    \begin{subfigure}[t]{\linewidth}
        \centering
        \includegraphics[width=\linewidth]
            {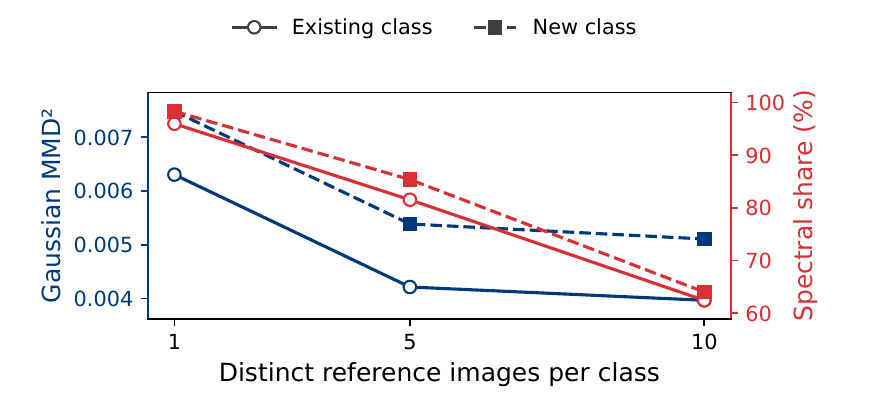}
        \caption{Fixed 20\% replacement.}
        \label{fig:cifar_dino_fixed}
    \end{subfigure}\par\smallskip

    \begin{subfigure}[t]{\linewidth}
        \centering
        \includegraphics[width=\linewidth]
            {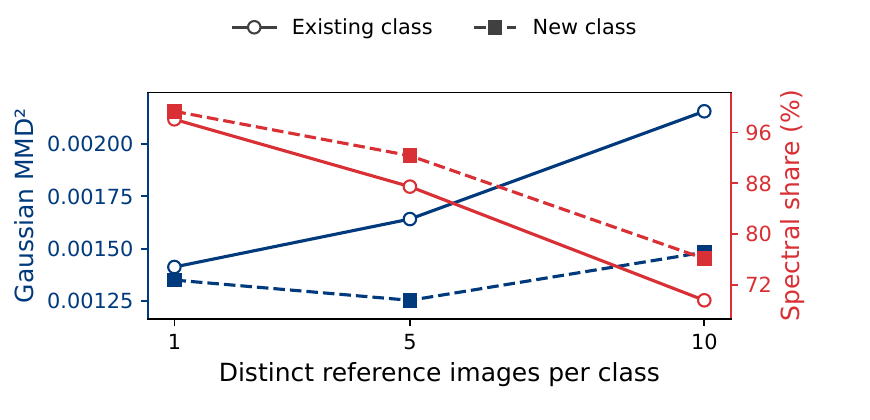}
        \caption{Matched BW distance.}
        \label{fig:cifar_dino_matched}
    \end{subfigure}

    \caption{\textbf{How reference diversity affects the measured shift.}
    (a) Example reference images from five classes (Ref.) and two sets of ten incoming images: cats, a class already present in the reference (Ext.), and airplanes, a class absent from the reference (New). 
    (b--c) Mean DINOv2 spectral share (red, right axis) falls with reference diversity under both constraints. For familiar-class additions, Gaussian MMD$^2$ (blue, left axis) falls under fixed replacement but rises under BW matching.}
    \label{fig:cifar_dino_fixed_matched}
\end{figure}

\paragraph{Familiar versus absent classes on CIFAR-10 \cite{krizhevsky_learning_2009}:} We ask whether spectral allocation helps distinguish the introduction of a previously absent class from new examples of a familiar class. Each reference contains five classes with one, five, or ten distinct images per class. From the same reference, we construct two endpoints: one receives ten fresh images from a single familiar class, and the other receives ten images from a single absent class (\cref{fig:cifar_replacement_samples}). In the fixed setting (F), we use the same replacement fraction for both interventions. In the matched setting (M), we choose the fraction separately for each endpoint to reach a common target squared BW distance in DINOv2 \cite{oquab_dinov2_2024} feature space; corresponding results using CLIP features appear in \cref{app:cifar_clip_results}. \cref{thm:spectral-trace-one} explains the purpose of this second control. For trace-one operators, the spectral share contains a contribution $D/4$ determined solely by the squared BW distance $D$. At our matched target \(D\approx.08105\), this contribution is approximately $2.03\%$ for both endpoints. Matching therefore holds the distance-dependent part of spectral share fixed: a gap between the two shares reflects a difference in the remaining, zero-trace displacement within reference eigenspaces.

\paragraph{Dependence on reference diversity:} With only one distinct image per reference class, mean spectral share exceeds $95\%$ for both interventions under both constraints (\cref{fig:cifar_dino_fixed} and~\cref{fig:cifar_dino_matched}). As the number of distinct images per class increases from one to five to ten, mean share falls at each step in all four comparisons. Spectral share is also lower with ten reference images per class than with one under a four-dimensional PCA projection in which all unregularized reference operators have full rank (\cref{app:cifar_pca4}). The trend therefore persists when reference rank is held fixed, although this check does not separate semantic coverage from estimation effects associated with more distinct reference images.

At matched BW distance, absent-class introduction has a higher mean spectral share than familiar-class replacement at every tested reference size in DINOv2. The gap widens from $1.3$ to $4.9$ to $6.6$ percentage points as the number of reference images per class increases from one to five to ten. By \cref{thm:spectral-trace-one}, the distance-dependent contribution is essentially identical for the two interventions ($D/4\approx 2.03\%$). The larger spectral share for absent-class introduction therefore reflects greater allocation to zero-trace displacement within the reference eigenspaces at the same overall shift magnitude.

\paragraph{Comparison with kernel discrepancy and diversity:} We compute Gaussian-kernel MMD$^2$ using the same normalized
DINOv2 features and sample weights as the BW decomposition. This follows the kernel-discrepancy approach of CMMD \cite{jayasumana_rethinking_2024}, using DINOv2 rather than CLIP \cite{radford_learning_2021} features. Under fixed replacement, mean MMD$^2$ decreases with reference size for both interventions (\cref{fig:cifar_dino_fixed}). Under BW matching, it increases for familiar-class additions and is nonmonotonic for absent-class additions (\cref{fig:cifar_dino_matched}): the mean falls from $0.00135$ to $0.00125$ before rising to $0.00148$. Matching BW distance therefore does not fix Gaussian-kernel MMD$^2$, while spectral share describes how the BW displacement is allocated within and across reference eigenspaces. Vendi and spectral share also answer different questions. Endpoint Vendi measures the effective rank of a collection's probability-weighted similarity matrix \cite{friedman_vendi_2023}, while spectral share measures how displacement from a specified reference is allocated. Across held-out matched comparisons, their Pearson correlations are $0.15$, $0.44$, and $0.58$ at one, five, and ten distinct reference images per class, respectively.

\paragraph{Incremental prediction:} We fit ridge-regularized logistic regression models to estimate the probability that the incoming class was absent from the reference. We select the regularization strength by five-fold block cross-validation. The baseline includes reference and endpoint Vendi scores, squared BW distance, the number of distinct reference images, replacement-weighted mean nearest-neighbor ($k=1$) and centroid distances, Gaussian MMD$^2$ on normalized DINOv2 and CLIP features, and CMMD \cite{jayasumana_rethinking_2024}, computed using the official implementation and scaled by the squared replacement fraction. We log-transform the Vendi scores, squared BW distance, reference image count, and kernel discrepancies, and include squared terms and pairwise products of the baseline predictors. Under BW matching, we match distances within a numerical tolerance and use the common target squared BW distance in the regression. The augmented model adds spectral share and its products with log reference image count and log reference effective rank. We evaluate both models on 30 held-out, image-disjoint blocks, each constructed from a pool of 200 images. We use the Brier score~\cite{brier_verification_1950}, the mean squared error between predicted probabilities and binary class-absence labels, where lower scores indicate more accurate probability estimates.

\begin{figure}[t]
    \centering
    \includegraphics[width=\linewidth]
        {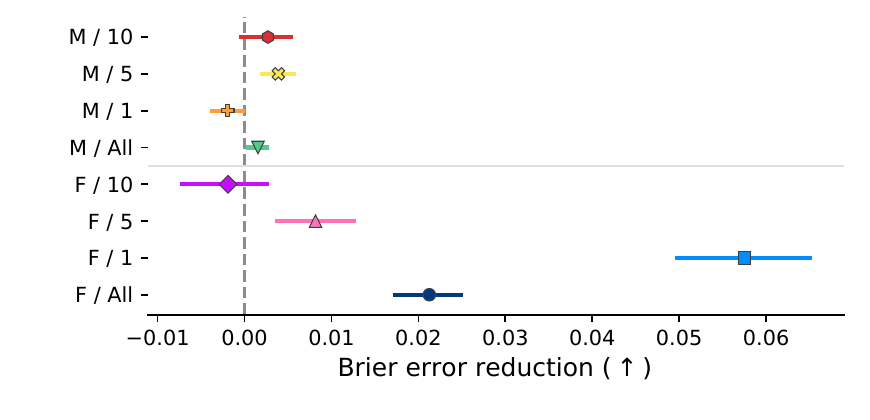}

    \caption{\textbf{Predictive gain from spectral allocation in DINOv2.}
Points show the reduction in held-out Brier error after adding spectral allocation to the baseline; positive values favor allocation. F denotes fixed replacement, pooling replacement fractions of $10\%$, $20\%$, and $40\%$; M denotes matched BW distance. Labels 1, 5, and 10 indicate distinct reference images per class; ``All'' pools these reference sizes. Bars show 97.5\% paired block-bootstrap intervals. For F/All and M/All, this level gives a Bonferroni correction for the two pooled comparisons within DINOv2. Reference-specific comparisons are descriptive.}
\label{fig:dino_brier_by_reference}
\end{figure}

Under fixed replacement, pooling the three reference sizes and the $10\%$, $20\%$, and $40\%$ replacement fractions, adding spectral share and its interaction terms lowers held-out Brier error from $0.1284$ to $0.1072$, an absolute reduction of $0.0213$ (approximately $16.6\%$ relative; F/All in \cref{fig:dino_brier_by_reference}). Descriptively, the gain is largest when the reference contains one distinct image per class: the absolute error reduction is $0.0575$ at one image and $0.0082$ at five, whereas error increases by $0.0019$ at ten. Under matched BW distance, pooled error falls from $0.0997$ to $0.0981$, a $1.6\%$ relative reduction (M/All in \cref{fig:dino_brier_by_reference}). With an exploratory Bonferroni correction \cite{dunn_multiple_1961} across F/All and M/All in DINOv2, the matched-BW reduction has a 97.5\% paired block-bootstrap confidence interval of $[0.000142,\,0.002820]$, excluding zero. Thus, the interventions differed in mean spectral share at matched BW distance, while the largest predictive gain occurred under fixed replacement with one reference image per class.

\section{Diagnosing Feature Shift under Classifier-Free Guidance}
\label{sec:stable_diffusion_exp}
We ask when CFG \cite{ho_classifier-free_2021} changes the allocation of displacement relative to prompt conditioning alone. The spectral shares of the CFG and prompt-only branches begin to diverge before requested categories are widely detected (\cref{fig:cfg_switch_timeline}). In contrast to the matched comparison in \cref{sec:controlled_experiments}, where BW distance is approximately fixed, the denoising experiment tracks a BW distance from the fixed reference that varies across updates. We therefore track spectral energy above its distance-dependent lower bound to quantify how much within-eigenspace energy exceeds the minimum implied by the observed distance. We then test whether applying CFG during the first or last five updates of the middle-noise interval affects final requested-category detection.

\paragraph{Separating prompt introduction from CFG:}
We use SDXL Base~1.0 \cite{podell_sdxl_2024} with 32 seeds per prompt on 40 two-object GenEval prompts \cite{ghosh_geneval_2023}. These prompts let us test how CFG timing affects the fraction of requested categories detected in each image. Following the limited-interval guidance approach of \citet{kynkaanniemi_applying_2024}, we sample SDXL using
32 deterministic Heun steps. We index denoising states by the number of completed updates, i.e., \(t=0\) denotes the initial noisy latent. Updates~11--26 fall in the middle-noise interval ($0.28<\sigma\leq5.42$, measured at the start of each update).

For each prompt--seed pair, the first ten updates use the empty prompt. All three branches start from the same latent at $t=10$. \emph{Null} retains the empty prompt. \emph{Prompt} introduces the target prompt at guidance scale $g=1$ for all remaining updates. \emph{CFG} introduces the same target prompt at $g=16$ for updates~11--26, then uses $g=1$ for the remaining updates. At states~0--31, we use the same empty prompt to obtain a clean-image estimate from each branch's current latent. We use these estimates for feature extraction and object detection. At state~32, we decode the final latent directly. For each prompt, branch, and state, the features of the 32 images define a regularized, trace-one uncentered second-moment operator. All three branches use the same fixed reference operator, formed from that prompt's 32 state-0 estimates. We use RT-DETR \citep{zhao_detrs_2024} at a confidence threshold of $0.5$ and score each requested category separately. Detecting one of the two categories gives the image a score of $1/2$. We report SDXL results in DINOv2 \cite{oquab_dinov2_2024} space here. \Cref{app:cfg_additional} provides CLIP \cite{radford_learning_2021} counterparts, sampling details, and cross-model results with FLUX.2 [klein] Base 4B \citep{black_forest_labs_flux2_2026}.

\paragraph{Spectral allocation at guidance onset:}
At the first guided update, from $t=10$ to $t=11$, mean spectral share rose by $5.51$ percentage points under CFG, compared with $1.04$ under Prompt and $1.55$ under Null. The increase was larger under CFG than under Prompt for all 40 prompts. At $t=11$, the detector recognized only $4.3\%$ of requested categories under CFG. By $t=14$, that fraction had reached $50.0\%$, compared with $1.3\%$ under Prompt (\cref{fig:cfg_switch_timeline}). The change in collection geometry thus preceded widespread detection.

Interpreting this rise in spectral share requires accounting for the change in squared BW distance. At $t=11$, the mean squared BW distance $D$ from the reference increased by $0.2500$ under CFG, versus $0.0522$ under Prompt. This matters since, by \cref{thm:spectral-trace-one}, the spectral share of a trace-one operator is bounded below by $D/4$. Under CFG, this floor rose by $6.25$ percentage points, slightly more than the observed $5.51$-point rise in spectral share. The share remained above its floor, but the gap narrowed by $0.74$ points. A smaller gap in \emph{share} does not necessarily indicate a smaller \emph{amount}
of spectral displacement, since the total distance also grew. 

\begin{figure}[t]
    \centering
    \begin{subfigure}[t]{\linewidth}
        \centering
        \includegraphics[width=\linewidth]
            {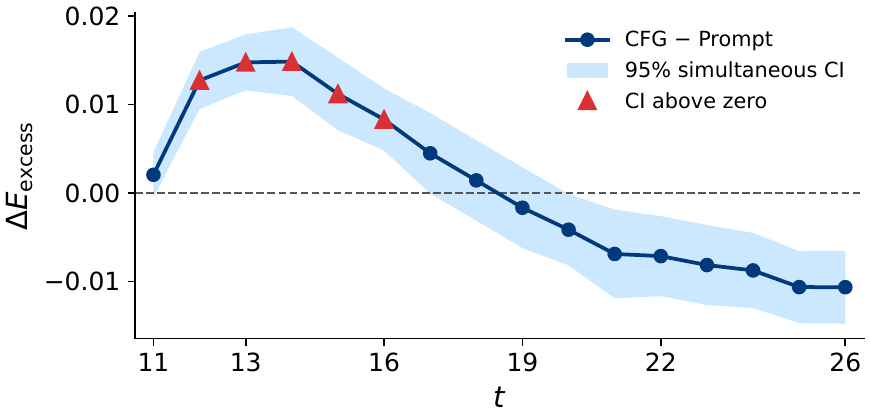}
        \caption{CFG-minus-Prompt excess spectral energy.}
        \label{fig:cfg_guided_excess_band}
    \end{subfigure}\par\smallskip
    \begin{subfigure}[t]{\linewidth}
        \centering
        \includegraphics[width=\linewidth]
            {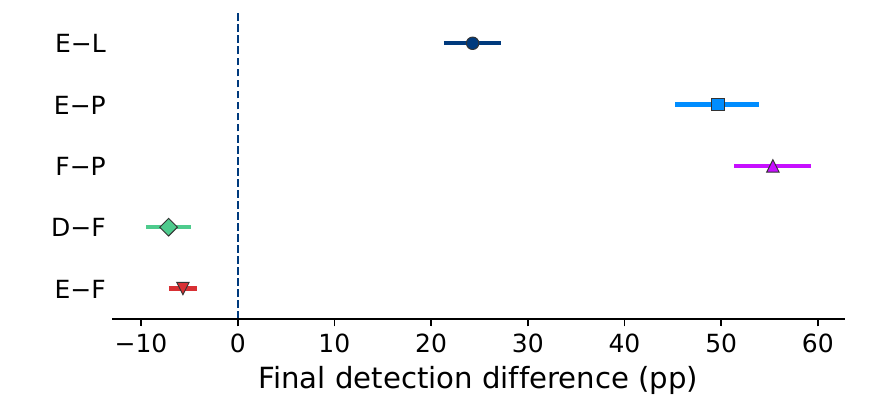}
        \caption{Paired differences in final detection.}
        \label{fig:cfg_schedule_detection}
    \end{subfigure}
    \caption{\textbf{Using spectral allocation to motivate a
    guidance-schedule test.}
    (a) $\Delta E_{\mathrm{excess}}$ is the mean CFG-minus-Prompt difference in excess spectral energy, $E_{\mathrm{excess}}=E_{\mathrm{spec}}-D^2/4$, where $D$ is squared BW distance. Shading shows the 95\% simultaneous band over guided updates; red triangles indicate updates where the simultaneous band lies entirely above zero.
    (b) Each row shows final detection under the first schedule minus that under the second, in percentage points. E guides at updates 11--15; L at 22--26; F at 11--26 (the full middle-noise interval); and D at 16--26 (delayed guidance). P uses prompt conditioning alone. Bars show paired 95\% prompt-family bootstrap intervals.}
    \label{fig:cfg_diagnostic_schedule}
\end{figure}

We thus compute excess spectral energy, $E_{\mathrm{spec}}-D^2/4$, for each branch and compare CFG with Prompt within each prompt. The mean difference was largest at $t=14$, reaching approximately $0.0148$ in squared BW units (\cref{fig:cfg_guided_excess_band}). A 95\% simultaneous prompt-family bootstrap band over all 16 guided updates lay entirely above zero at $t=12$--$16$, the second through sixth guided updates, and entirely below zero
at $t=20$--$26$.

\paragraph{Testing where to apply guidance:}
Limiting CFG to fewer updates can reduce inference cost \cite{kynkaanniemi_applying_2024}. Motivated by the early excess-energy response, we compared guidance during the first five (E, updates~11--15) and last five (L, updates~22--26) updates of the middle-noise interval, holding prompts, seeds, and guidance scale fixed. Final requested-category detection was $68.2\%$ under E and $43.9\%$ under L, a paired gain of $24.3$ percentage points (95\% prompt-family bootstrap CI $[21.3,\,27.2]$; \cref{fig:cfg_schedule_detection}). Early guidance yielded higher detection for all 40 prompts. A cross-model check with FLUX also shows higher final detection under early than late guidance (\cref{app:flux_guidance}).

Relative to prompt conditioning alone, early and full-interval guidance improved final detection by $49.7$ and $55.4$ percentage points, respectively. Early guidance thus recovered about $90\%$ of the full-interval gain, although final detection was $5.7$ points lower than under full-interval guidance. Early guidance did not meet our noninferiority criterion \citep{hahn_understanding_2012} since the one-sided 95\% lower bound for $\mathrm{E}-\mathrm{F}$ was $-6.9$ points, below the prespecified $-5$-point margin. Across prompts, the mean early CFG-minus-Prompt excess-energy contrast was only weakly associated with the early-minus-late detection gain (Spearman $\rho=0.18$). Taken together, these results suggest that the early response helped identify when to apply a five-update guidance budget, although its magnitude gave little indication of which prompts benefited most from applying guidance early.

\paragraph{Numerical robustness:}
For the three SDXL prompts tested, scaling the grouping tolerance by $10^{-1}$, $10^0$, or $10^1$ left the eigenvalue groups and hence both projections unchanged. Rotating the basis within repeated eigenspaces changed the component energies only at roundoff precision, consistent with \cref{prop:organizational-faithfulness}. Across all tested shrinkage values, the CFG-minus-Prompt excess-energy contrast varied in magnitude but had pointwise 95\% confidence intervals entirely above zero at updates~12--15. At update~11, the mean contrast was positive for every shrinkage value, but its interval included zero at the two smallest values. \cref{app:sdxl_stability} reports more details on robustness tests.
\section{Conclusion}
Our decomposition allocates squared BW distance exactly between displacement within and across reference eigenspaces. Controlled Gaussian mixtures illustrate how sharply this allocation can differ at comparable distances. For trace-one operators, squared BW distance determines a lower bound on spectral energy. Matching distance thus holds the bound fixed. On CIFAR-10, mean spectral share fell with increasing reference diversity under both interventions and both controls. At matched BW distance, absent-class replacements had higher mean share than familiar-class replacements at every tested reference size. Adding spectral share and its interactions improved held-out prediction most under fixed replacement, with only a small gain at matched distance. In SDXL, squared BW distance changes across denoising updates. Tracking spectral energy above the corresponding bound revealed an early response to CFG that motivated an equal-budget timing test. Guiding during the first five updates of the middle-noise interval improved final requested-category detection over guiding during the last five, although full-interval guidance achieved the highest detection rate. Across these experiments, the allocation helped distinguish shifts of similar magnitude and identify a CFG window worth testing.

\paragraph{Limitations:}
The allocation depends on the feature encoder and reference collection. We tested eigenvalue-grouping tolerance and basis rotation on three SDXL prompts. The early excess-energy response persisted across all tested shrinkage values, although its magnitude varied with shrinkage. Our subsampling experiments measured how operator estimates changed with sample size. For a fixed intervention, how much spectral share varies across independent reference samples of the same size remains open.
\section*{LLM Usage Statement}

We used OpenAI's ChatGPT to edit the manuscript, identify relevant literature, and assist with the theory, experiments, and analyses. For \cref{sec:geometric_diagnostic} and the associated proofs, we provided the problem setup and initial mathematical ideas. In a series of exchanges, ChatGPT helped formulate the theoretical results and generated draft proof arguments, which we reviewed and refined. We also used ChatGPT to refine experimental methods and analyses we had proposed. ChatGPT generated initial drafts of the experimental and analysis code, which we then revised. All use was interactive, and we used no autonomous agents. We reviewed the text, proofs, and code and take responsibility for the final manuscript. We developed the research ideas and set the direction of the work.
{
    \small
    \bibliographystyle{unsrtnat}
    \bibliography{main}
}

\clearpage
\onecolumn
\appendix
\startcontents[appendix]

\section*{Appendix contents}
\printcontents[appendix]{}{1}{\setcounter{tocdepth}{2}}

\twocolumn

\section{Operator construction and related metrics}
\label{app:methodology}

\subsection{Feature normalization and operator construction}
\label{app:operator_construction}

Let \(f:\mathcal X\to\mathbb R^d\) denote the raw feature map. We define the pointwise \(\ell_2\)-normalized feature representation
\begin{equation}
\varphi(x)
=
\frac{f(x)}
{\max\{\|f(x)\|_2,\varepsilon_{\mathrm{norm}}\}},
\end{equation}
where \(\varepsilon_{\mathrm{norm}}>0\) is a small numerical safeguard.

At the population level, define
\begin{equation}
\label{eq:population-feature-second-moment}
M_P
:=
\mathbb E_{X\sim P}
\left[
\varphi(X)\varphi(X)^\top
\right],
\end{equation}
with the corresponding trace-normalized operator
\begin{equation}
\label{eq:population-density-operator}
\rho(P)
:=
\frac{M_P}
     {\operatorname{tr}(M_P)}.
\end{equation}
The entries of \(\rho(P)\) are normalized joint feature-activation moments:
\begin{equation}
\label{eq:feature-coactivation}
\rho(P)_{jk}
=
\frac{
    \mathbb E_{X\sim P}
    [\varphi_j(X)\varphi_k(X)]
}{
    \mathbb E_{X\sim P}
    \|\varphi(X)\|_2^2
}.
\end{equation}

If all raw feature vectors have norm at least $\varepsilon_{\mathrm{norm}}$, then the normalized features have unit norm and $\operatorname{tr}(M_P)=1$. Features below this threshold have norms smaller than one after normalization. We therefore retain explicit trace normalization, provided $\operatorname{tr}(M_P)>0$.

\subsection{Relation to FID and Gaussian Wasserstein geometry}
\label{app:fid_relation}

Our use of Bures--Wasserstein geometry is closely related to, but conceptually distinct from, the covariance term appearing in the Fr\'{e}chet Inception Distance (FID). We clarify this relationship here.

Let \(P\) and \(Q\) denote two feature distributions with means \(\mu_P,\mu_Q\) and centered covariance matrices
\(\Sigma_P,\Sigma_Q\). FID evaluates
\begin{equation}
\label{eq:fid_gaussian}
\begin{aligned}
\mathrm{FID}(P,Q)
&= \|\mu_P-\mu_Q\|_2^2
   + \operatorname{tr}(\Sigma_P+\Sigma_Q) \\
&\quad - 2\,\operatorname{tr}\!\left[
   \bigl(\Sigma_P^{1/2}\Sigma_Q\Sigma_P^{1/2}\bigr)^{1/2}
   \right].
\end{aligned}
\end{equation}

Under the Gaussian approximation used by FID, this expression is the squared \(2\)-Wasserstein distance between the fitted Gaussian feature distributions \citep{dowson_frechet_1982,heusel_gans_2017},
\begin{equation}
\mathcal N(\mu_P,\Sigma_P)
\quad\text{and}\quad
\mathcal N(\mu_Q,\Sigma_Q).
\end{equation}

For positive-semidefinite matrices \(A\) and \(B\), the squared Bures--Wasserstein distance is
\begin{equation}
\label{eq:bures_fid_relation}
\begin{aligned}
d_{\mathrm{BW}}^2(A,B)
&= \operatorname{tr}(A) + \operatorname{tr}(B) \\
&\quad - 2\operatorname{tr}\!\left[
    \bigl(A^{1/2} B A^{1/2}\bigr)^{1/2}
\right].
\end{aligned}
\end{equation}
Hence, when \(A=\Sigma_P\) and \(B=\Sigma_Q\), the covariance contribution in \cref{eq:fid_gaussian} is exactly \(d_{\mathrm{BW}}^2(\Sigma_P,\Sigma_Q)\)
\citep{bhatia_bureswasserstein_2019}. 

Equivalently, for centered Gaussian distributions,
\begin{equation}
W_2^2\!\left(
\mathcal N(0,A),
\mathcal N(0,B)
\right)
=
d_{\mathrm{BW}}^2(A,B).
\end{equation}
This identity is the optimal-transport origin of the Bures--Wasserstein terminology: on covariance matrices of centered Gaussian
distributions, the matrix distance above coincides with the $L^2$-Wasserstein distance \citep{van_oostrum_bureswasserstein_2022}.

BW distance compares positive-semidefinite operators without requiring Gaussian feature distributions. Its interpretation as the Wasserstein distance between the underlying distributions, however, generally requires Gaussianity.

The comparison captures only the information in the chosen operators. In particular, identical covariance matrices give zero covariance-level BW distance even when the feature distributions differ.

Our empirical construction additionally differs from the centered covariance used in FID. Given feature vectors \(z_1,\ldots,z_n\), we form the uncentered second-moment operator
\begin{equation}
S
=
\frac{1}{n}
\sum_{i=1}^{n}
z_i z_i^\top.
\end{equation}
At the population level,
\begin{equation}
S
=
\Sigma+\mu\mu^\top,
\end{equation}
so the operator retains mean-direction structure within the second-moment representation rather than separating it into an independent Euclidean mean term as in FID. This changes the operator whose geometry is analyzed, but does not change the fact that the construction remains second-order.

\subsection{Feature encoders and preprocessing}
\label{app:feature_encoders}

\paragraph{Synthetic feature representation:}
We use eight Gaussian RBF features centered at the reference mixture means, with bandwidth $\sigma_{\mathrm{RBF}}=0.50$:
\begin{equation}
\begin{aligned}
r_j(x)
&= \exp\!\left(
    -\frac{\|x-\mu_j\|_2^2}{2\sigma_{\mathrm{RBF}}^2}
\right), \\
\phi(x)
&= \frac{r(x)}{\max\{\|r(x)\|_2,10^{-12}\}}.
\end{aligned}
\end{equation}
Centers remain fixed across candidate distributions. We form $M_Q=\mathbb{E}_{X\sim Q}[\phi(X)\phi(X)^\top]$ and use
\begin{equation}
\rho_\delta(Q)
=(1-\delta)\frac{M_Q}{\operatorname{tr}(M_Q)}
+\delta\frac{I_8}{8}.
\end{equation}

\paragraph{Pretrained image features:}
\label{app:image_features}
For image distributions, the hand-designed RBF map is replaced by a frozen pretrained image encoder
\begin{equation}
f:
\mathcal X_{\mathrm{image}}
\to
\mathbb R^{d_f}.
\end{equation}
We use DINOv2 ViT-S/14 \citep{oquab_dinov2_2024} and CLIP ViT-L/14@336px \citep{radford_learning_2021}. Specifically, we use DINOv2's final classification (CLS) token as the image-level feature vector and CLIP's final projected image embedding. For each representation space, the same frozen encoder and preprocessing are applied to all reference and comparison images. For both encoders, we apply the normalization in \cref{app:operator_construction} with $\varepsilon_{\mathrm{norm}}=10^{-12}$ and construct the empirical operators as in \cref{subsec:feature_distribution_operators}.

The following preprocessing describes the features used for BW operators and the normalized-feature Gaussian MMD$^2$ comparisons. All encoder inputs are RGB images. For DINOv2, we resize directly to $224\times224$ pixels using PIL bicubic interpolation, without cropping. Pixel values are scaled to $[0,1]$ and normalized channel-wise using RGB means $(0.485,0.456,0.406)$ and standard deviations $(0.229,0.224,0.225)$.

For CLIP ViT-L/14@336px, we use the checkpoint's \texttt{CLIPImageProcessor}: bicubic resizing of the shorter side to 336 pixels, followed by a $336\times336$ center crop. Pixel values are scaled to $[0,1]$ and normalized using RGB means $(0.48145466,0.45782750,0.40821073)$ and standard deviations $(0.26862954,0.26130258,0.27577711)$. These pixel-level transformations precede the embedding normalization defined above.

\paragraph{Official CMMD preprocessing:}
The official CMMD baseline \cite{jayasumana_rethinking_2024} uses its own image preprocessing and CLIP embedding pipeline, rather than substituting the pointwise-normalized embeddings defined above. We retain the official pipeline and kernel settings; the endpoint weighting and kernel discrepancies are described in \cref{app:cifar_prediction}.
\section{Numerical implementation and shrinkage calibration}
\label{app:operator_regularization}

\subsection{Regularization and positive definiteness}
\label{app:shrinkage_admissibility}

The BW logarithm and tangent metric are defined on positive-definite operators \citep{bhatia_bureswasserstein_2019}. Empirical second-moment operators can be singular, so we apply isotropic shrinkage to the trace-normalized estimates before computing the decomposition. \cref{lem:regularized_operator_bounds} shows that the regularized operators are positive definite for every $0<\delta<1$.

For fixed $d$ and $\delta$, the same lemma keeps the operators uniformly away from singularity and bounds the BW tangent norm relative to the Frobenius norm on $\mathcal D_{d,\delta}$. These bounds do not ensure that empirical eigenspaces are stable when distinct eigenvalues are close; that would require an additional eigengap assumption. Likewise, the finite-sample bound in \cref{lem:operator_mse} concerns error in the operator itself, not in its spectral share.

\subsection{Independent selection of the shrinkage strength}
\label{app:shrinkage_selection}

Shrinkage is widely used to improve covariance estimates from finite samples \citep{ledoit_well-conditioned_2004,tong_linear_2018,raninen_coupled_2021}. Here we use it to control numerical conditioning in the BW computations. We first select the smallest positive grid candidate satisfying a fixed condition-number threshold on independent calibration data. We then multiply this minimum by ten as a common numerical safety margin, improving conditioning at the cost of a larger contribution from the isotropic target.

\paragraph{Calibration rule:}
Let $j$ index calibration collections, with $n_{\mathrm{cal},j}$ unit-normalized feature vectors in collection $j$. We define
\begin{equation}
\hat{\rho}_{\mathrm{cal},j}
=
\frac{1}{n_{\mathrm{cal},j}}
\sum_{i\in\mathrm{cal}_j}
\phi_i\phi_i^\top
\end{equation}
For Gaussian mixtures and CIFAR-10 \cite{krizhevsky_learning_2009}, we construct one operator from all calibration samples for each feature map. For SDXL \cite{podell_sdxl_2024} and FLUX \cite{black_forest_labs_flux2_2026}, we construct 16 separate calibration reference operators per encoder. For each positive candidate $\delta$, we form
\begin{equation}\label{eq:isotropic_shrinkage}
\hat{\rho}_{\mathrm{cal},j,\delta}
=
(1-\delta)\hat{\rho}_{\mathrm{cal},j}
+
\delta\frac{I_d}{d}.
\end{equation}
We accept a candidate when every calibration operator satisfies the conditioning threshold:
\begin{equation}
\begin{aligned}
&\kappa(\hat{\rho}_{\mathrm{cal},j,\delta})=\frac{
\lambda_{\max}(\hat{\rho}_{\mathrm{cal},j,\delta})
}{
\lambda_{\min}(\hat{\rho}_{\mathrm{cal},j,\delta})
},\\
&\max_j\kappa(\hat{\rho}_{\mathrm{cal},j,\delta})\le \kappa_{\mathrm{th}}.
\end{aligned}
\label{eq:delta_conditioning}
\end{equation}
Here $\kappa(\cdot)$ is the spectral condition number, which measures sensitivity to relative perturbations \citep{trefethen_numerical_1997}. For symmetric positive-definite matrices, the singular values equal the eigenvalues, which gives the ratio above.

We set
\begin{equation}
\kappa_{\mathrm{th}}
=
\epsilon_{\mathrm{fp}}^{-1/2},
\end{equation}
where $\epsilon_{\mathrm{fp}}$ is double-precision floating-point epsilon. For float64, this gives $\kappa_{\mathrm{th}}\approx6.7\times10^7$. The choice follows the usual rule of thumb that a condition number $\kappa$ can cost roughly $\log_{10}\kappa$ decimal digits of accuracy \citep{trefethen_numerical_1997}. The threshold lies halfway between $1$ and $\epsilon_{\mathrm{fp}}^{-1}$ on a logarithmic scale, allocating about half of the nominal precision to conditioning. This is a numerical criterion, rather than an error bound for the full BW computation.

For a fixed grid $\mathcal D_+$ of positive candidates, we select
\begin{equation}
\delta_{\min}
=
\min
\left\{
\delta\in\mathcal{D}_{+}:
\max_j\kappa(\hat{\rho}_{\mathrm{cal},j,\delta})
\le
\kappa_{\mathrm{th}}
\right\}.
\label{eq:delta_selection}
\end{equation}
Across all experiments, the analysis value is
\begin{equation}
\delta_{\mathrm{used}}=10\delta_{\min}.
\label{eq:delta_used}
\end{equation}
All resulting analysis values satisfy $0<\delta_{\mathrm{used}}<1$. Both $\delta_{\min}$ and $\delta_{\mathrm{used}}$ ensure positive definiteness; the larger value increases the smallest eigenvalues and improves conditioning.

Calibration is performed separately for each experimental setting and feature representation, and $\delta_{\mathrm{used}}$ is held fixed across the corresponding comparisons. Neither the component energies nor the subsequent sensitivity analyses are used to choose it. Conditioning and numerical checks on experimental operators are reported separately and do not revise the selected value. The analyses reported in the main paper use this common tenfold margin, which was not prespecified for the initial runs.

\subsection{Calibration sets and selected values}
\label{app:calibration_sets}

\paragraph{Gaussian-mixture calibration:}
\label{app:shrinkage_selection_controlled}
We draw an independent calibration sample of $n_{\mathrm{cal}}=4000$ observations from the reference eight-mode Gaussian mixture and apply the same normalized RBF feature map used in the experiment. Calibration is completed before generating any coverage-imbalance, mode-merging, or spurious-mode candidate. Calibration and experimental draws use independent random streams.

The positive candidate grid spans $10^{-8}$ to $5\times10^{-1}$. The unregularized calibration operator has condition number $3.34$, already below $\kappa_{\mathrm{th}}$. The smallest positive grid candidate therefore passes, giving $\delta_{\min}=10^{-8}$ and $\delta_{\mathrm{used}}=10^{-7}$. Following the tenfold margin in \cref{app:shrinkage_selection}, we use $\delta_{\mathrm{used}}=10^{-7}$ for the reference and all three intervention operators. Repeating
calibration on 20 subsets at each of 1000, 2000, and 4000 observations gives the same minimum throughout.

\paragraph{Calibration for the image experiments:}
\label{app:shrinkage_selection_sdxl}

For CIFAR-10, calibration uses 4,000 images from the official test split, disjoint from the training images used in the experiments. For each encoder, we construct one operator from all calibration images and select $\delta_{\min}$ from
\begin{equation}
\begin{aligned}
\mathcal D_+=
\{10^{-k},3\times10^{-k}:k=3,\ldots,8\}&\\
\cup\{0.01,0.03,0.05,0.1,0.2,0.5\}&.
\end{aligned}
\end{equation}

For SDXL, calibration uses 16 separate prompts with 16 seeds each, disjoint from the 40 experimental prompts. We construct one state-0 operator per calibration prompt and require all 16 operators to satisfy the conditioning threshold. This calibration remains fixed for the main experiment, which uses 32 seeds per prompt. The candidate grid is
\begin{equation}
\label{eq:diffusion_delta_grid}
\begin{aligned}
\mathcal D_+ ={}&
\{10^{-k},\,3\times10^{-k}:k=3,\ldots,8\}\\
&{}\cup\{0.01,0.02,0.03,0.05,0.075,0.1,\\
&\hspace{2.8em}0.15,0.2,0.3,0.5\},
\end{aligned}
\end{equation}
which samples small shrinkage values at approximately logarithmic intervals and includes larger candidates up to $0.5$. DINOv2 and CLIP are calibrated separately using the same calibration images. 

For FLUX, calibration uses 16 independent groups of 16 state-0 readouts. Each calibration image is an empty-text state-0 readout from an independently sampled latent. The 256 calibration seeds are disjoint from the experimental seeds. We use the candidate grid in \cref{eq:diffusion_delta_grid} and require all 16 calibration operators to satisfy the conditioning threshold. DINOv2 and CLIP are calibrated separately using the same calibration images. The selected minimum
and analysis values are listed in \cref{tab:calibrated_shrinkage}.

In all three image experiments, we use $\delta_{\mathrm{used}}=10\delta_{\min}$ and hold it fixed across the corresponding comparisons. Evaluation outcomes are not used to select shrinkage.

\begin{table}[t]
\centering
\small
\begin{tabular}{@{}llcc@{}}
\toprule
Experiment & Encoder & $\delta_{\min}$ & $\delta_{\mathrm{used}}$ \\
\midrule
CIFAR-10 & DINOv2 & $10^{-6}$ & $10^{-5}$ \\
CIFAR-10 & CLIP   & $10^{-5}$ & $10^{-4}$ \\
SDXL     & DINOv2 & $10^{-5}$ & $10^{-4}$ \\
SDXL     & CLIP   & $3\times10^{-5}$ & $3\times10^{-4}$ \\
FLUX     & DINOv2 & $10^{-5}$ & $10^{-4}$ \\
FLUX     & CLIP   & $3\times10^{-5}$ & $3\times10^{-4}$ \\
\bottomrule
\end{tabular}
\caption{Independent shrinkage calibration for the CIFAR-10, SDXL, and FLUX analyses. Each experiment and encoder is calibrated separately. The analysis value is ten times the smallest passing grid candidate.}
\label{tab:calibrated_shrinkage}
\end{table}

\subsection{BW computation, eigenvalue grouping, and numerical checks}
\label{app:robustness}
\label{app:eigenvalue_grouping}

\paragraph{Numerical BW computation and equivalence:}
We explain how the factor-based implementation evaluates the distance, transport map, and component energies defined in the main text. Let $A,B\succ0$, and choose invertible factors satisfying $A=L_A L_A^\top$ and $B=L_B L_B^\top$. SDXL and FLUX use Cholesky factors. CIFAR-10 instead performs the calculation in the reference eigenbasis, using a diagonal reference factor and a Cholesky factor for the endpoint. These choices give the same BW quantities in exact arithmetic. Cholesky factorization, singular value decomposition, and linear solves are standard numerical linear-algebra operations \citep{trefethen_numerical_1997}.

\emph{Factor alignment:}
A factor is not unique: for any orthogonal matrix $R$, $(L_B R)(L_B R)^\top=B$. Thus, right multiplication by $R$ changes the factor representation without changing the operator. The BW distance admits the equivalent factor-alignment formulation \citep{bhatia_bureswasserstein_2019}
\begin{equation}
d_{\mathrm{BW}}^2(A,B)
=
\min_{R^\top R=I}
\|L_B R-L_A\|_F^2.
\end{equation}
Expanding the objective gives
\begin{equation}
\|L_B R-L_A\|_F^2
=
\operatorname{tr}(A)+\operatorname{tr}(B)
-
2\operatorname{tr}(L_A^\top L_B R).
\end{equation}
The minimization therefore amounts to maximizing the final trace. If $L_A^\top L_B=U_s\Sigma V_s^\top$ is a singular value decomposition, its maximum is $\operatorname{tr}(\Sigma)$, attained at $R_\star=V_sU_s^\top$. We consequently compute
\begin{equation}
\begin{aligned}
L_A^\top L_B &= U_s\Sigma V_s^\top,\\
R_\star &= V_sU_s^\top,\\
\Delta &= L_B R_\star-L_A,\\
d_{\mathrm{BW}}^2(A,B) &= \|\Delta\|_F^2.
\end{aligned}
\label{eq:bw_factor_computation}
\end{equation}

\emph{Recovering the transport map:}
To connect the aligned factors to the transport formula in the main text, define $K=L_B R_\star L_A^{-1}$. The alignment obtained by singular value decomposition gives
\begin{equation}
\begin{aligned}
L_A^\top K L_A
&=
U_s\Sigma U_s^\top
\succ0,\\
KAK^\top
&=
L_B R_\star R_\star^\top L_B^\top
=
B.
\end{aligned}
\end{equation}
The first identity gives $K=L_A^{-\top}(U_s\Sigma U_s^\top)L_A^{-1}$, so $K$ is symmetric positive definite. The second identity therefore becomes $KAK=B$. Setting $C=A^{1/2}KA^{1/2}\succ0$, we obtain $C^2=A^{1/2}BA^{1/2}$. Uniqueness of the positive-definite square root therefore yields
\begin{equation}
\begin{aligned}
K
&=
A^{-1/2}
\bigl(A^{1/2}BA^{1/2}\bigr)^{1/2}
A^{-1/2}\\
&=
T_{A\to B}.
\end{aligned}
\end{equation}
Hence the aligned factor difference satisfies
\begin{equation}
\begin{aligned}
\Delta &= (T_{A\to B}-I)L_A,\\
E &:= \Delta L_A^{-1} = T_{A\to B}-I.
\end{aligned}
\end{equation}
The SDXL and FLUX implementations obtain $E$ by solving $L_A^\top E^\top=\Delta^\top$, without explicitly forming $L_A^{-1}$. In exact arithmetic, $E$ is symmetric; the computed matrix is symmetrized to remove its antisymmetric numerical part.

\emph{Logarithmic displacement and total energy:}
The main-text logarithmic displacement is $V_{A\to B}=AE+EA$. Thus, for the Lyapunov operator $\mathcal L_A(X)=AX+XA$, we have $\mathcal L_A^{-1}(V_{A\to B})=E$. Substitution into the BW metric gives
\begin{equation}
\begin{aligned}
\|V_{A\to B}\|_A^2
&=
\frac12\operatorname{tr}(V_{A\to B}E)\\
&=
\operatorname{tr}(EAE)\\
&=
\|E L_A\|_F^2\\
&=
\|\Delta\|_F^2.
\end{aligned}
\end{equation}
This also recovers the square-root expression directly. Writing $T=T_{A\to B}$ and $C=(A^{1/2}BA^{1/2})^{1/2}$, the identities $TAT=B$ and $\operatorname{tr}(TA)=\operatorname{tr}(C)$ give
\begin{equation}
\begin{aligned}
\operatorname{tr}(EAE)
&=
\operatorname{tr}\bigl((T-I)A(T-I)\bigr)\\
&=
\operatorname{tr}(A)+\operatorname{tr}(B)
-
2\operatorname{tr}(C)\\
&=
d_{\mathrm{BW}}^2(A,B).
\end{aligned}
\end{equation}
Thus, the factor computation and the square-root formula evaluate the same distance in exact arithmetic.

\emph{Component energies:}
The SDXL and FLUX implementations retain $E$ rather than explicitly constructing $V_{A\to B}$. To see why this suffices, write $A=Q\Lambda Q^\top$, with
$\Lambda=\operatorname{diag}(\lambda_1,\ldots,\lambda_d)$, and let $Z=Q^\top E Q$. Then
\begin{equation}
\bigl(Q^\top V_{A\to B}Q\bigr)_{ij}
=
(\lambda_i+\lambda_j)Z_{ij}.
\end{equation}
Substituting these entries into the coordinate expression for the BW metric gives
\begin{equation}
\begin{aligned}
\|V_{A\to B}\|_A^2
&=
\frac12
\sum_{i,j}
\frac{
\bigl(Q^\top V_{A\to B}Q\bigr)_{ij}^{\,2}
}{
\lambda_i+\lambda_j
}\\
&=
\frac12
\sum_{i,j}
(\lambda_i+\lambda_j)Z_{ij}^{\,2}.
\end{aligned}
\label{eq:bw_energy_computation}
\end{equation}
Let $\mathcal I_{\mathrm{within}}$ contain the index pairs belonging to the same reference eigenspace. The spectral projection retains precisely these entries, while the organizational projection retains the complementary entries. The corresponding energies are therefore
\begin{equation}
\begin{aligned}
E_{\mathrm{spec}|A}(B)
&=
\frac12
\sum_{(i,j)\in\mathcal I_{\mathrm{within}}}
(\lambda_i+\lambda_j)Z_{ij}^{\,2},\\
E_{\mathrm{org}|A}(B)
&=
\frac12
\sum_{(i,j)\notin\mathcal I_{\mathrm{within}}}
(\lambda_i+\lambda_j)Z_{ij}^{\,2}.
\end{aligned}
\end{equation}
SDXL and FLUX evaluate these weighted within-block and across-block sums. CIFAR-10 computes the aligned factor difference directly in the reference eigenbasis,
where $W=Z\Lambda^{1/2}$, and sums $W_{ij}^2$ within and across the same blocks. These calculations are equivalent: $Z$ is symmetric and each index set is
closed under exchanging $i$ and $j$, so
\begin{equation}
\sum_{(i,j)\in\mathcal I} W_{ij}^2
=
\sum_{(i,j)\in\mathcal I}\lambda_j Z_{ij}^2
=
\frac12\sum_{(i,j)\in\mathcal I}
(\lambda_i+\lambda_j)Z_{ij}^2.
\end{equation}
Both implementations therefore evaluate the same component energies in exact arithmetic. The Gaussian-mixture experiment evaluates the matrix-square-root formulas directly.

\paragraph{Reduction to the sample span:}
For the image experiments, we use QR factorization to construct an orthonormal basis containing the reference and comparison feature vectors to reduce the matrix size, and perform the matrix calculations in that subspace. Outside it, both regularized operators equal $(\delta/d)I$, so their displacement is zero.
This reduction preserves the BW distance and both component energies without discarding observed feature directions. We retain the original feature dimension
$d$ in the shrinkage term $\delta I/d$ and do not renormalize the reduced operators.

We compute the eigendecomposition of the regularized reference operator in double precision and sort its eigenvalues in descending order. Adjacent eigenvalues
are assigned to the same group when
\begin{equation}
|\lambda_i-\lambda_{i+1}|
\leq
\tau_{\mathrm{abs}}
+
\tau_{\mathrm{rel}}
\max\{1,|\lambda_i|,|\lambda_{i+1}|\}.
\end{equation}
For the image experiments, we use $\tau_{\rm abs}=10^{-12}$ and $\tau_{\rm rel}=10^{-10}$. The Gaussian-mixture implementation uses $\tau_{\rm abs}=10^{-10}$ and $\tau_{\rm rel}=10^{-8}$. Groups are formed by joining consecutive eigenvalues satisfying this rule. For trace-one positive-definite operators, the threshold is \(1.01\times10^{-10}\) in the image experiments and \(1.01\times10^{-8}\) in the Gaussian-mixture experiment.

Each group is treated as a full subspace: the spectral projection retains the entire within-group block, including its off-diagonal entries, while the
organizational projection retains the between-group blocks. In particular, directions in the empirical nullspace all acquire eigenvalue $\delta/d$ after shrinkage. Thus, the repeated eigenvalue $\delta/d$ arising from the empirical nullspace is handled as one group. Rotating the basis within an exactly repeated
eigenspace leaves the allocation unchanged. Grouping numerically close but distinct eigenvalues is a numerical approximation to the exact eigenspace partition.

We check positive definiteness and numerical conditioning of the regularized operators and monitor the energy identity
\begin{equation}
\begin{aligned}
\left|
d_{\mathrm{BW}}^2(A,B)
-
E_{\mathrm{spec}}
-
E_{\mathrm{org}}
\right|&
\\
\leq 10^{-8}\max\{d_{\mathrm{BW}}^2(A,B),10^{-4}\}&.
\end{aligned}
\end{equation}
In the image experiments, we leave spectral share undefined when the total component energy is at most $10^{-10}$, since division by a near-zero total makes the ratio numerically unreliable. The experiment-specific sections report checks using tenfold smaller and larger grouping tolerances and random basis rotations within repeated eigenspaces.
\section{Controlled experiments}
\label{app:controlled}

\subsection{Gaussian mixture experiment}\label{app:gaussian_mixture}

\subsubsection{Reference construction and controlled perturbations}
\label{app:controlled_construction}
\label{app:controlled_perturbations}

We use eight mode-centered RBF features with bandwidth $\sigma_{\mathrm{RBF}}=0.50$, fixed before evaluating the perturbations. The reference is an eight-mode Gaussian mixture with centers equally spaced on a circle of radius $2.1$, component covariance $0.12^2I_2$, and weights
\begin{equation}
\begin{aligned}
p = (&0.20,\,0.17,\,0.15,\,0.13,\\
     &0.11,\,0.09,\,0.08,\,0.07).
\end{aligned}
\end{equation}
We use unequal weights such that the ideal reference operator $\operatorname{diag}(p_1,\ldots,p_8)$ has distinct eigenvalues. We estimate each conditional feature moment $C_k=\mathbb E[\varphi(X)\varphi(X)^\top\mid M=k]$ with 30,000 Gaussian draws and form $\widehat M_{\mathrm{ref}}=\sum_kp_k\widehat C_k$.
Common Gaussian draws are reused across component locations and severity levels. Reference and candidate operators are trace normalized and regularized with $\delta=10^{-7}$ (\cref{app:shrinkage_selection_controlled}); we denote them by $A$ and $B$. Endpoint visualizations use 4,000 separately generated
samples per condition.

At this bandwidth, samples from mode $k$ mainly activate the corresponding RBF coordinate, giving
\begin{equation}
\varphi(X)\approx e_k,
\qquad
\widehat M_{\mathrm{ref}}\approx\operatorname{diag}(p_1,\ldots,p_8).
\end{equation}
With sufficiently separated reference eigenvalues, the eigendirections also approximately align with the mode coordinates. Reweighting fixed modes then mainly changes diagonal structure, whereas activating several coordinates together can introduce cross-eigenspace coupling.

\paragraph{Matched distribution:}
We set $B=A$ using an exact copy of the estimated reference operator, so the discrepancy and both component energies vanish up to floating-point error.

\paragraph{Coverage imbalance:}
We keep all mode locations and covariances fixed and interpolate the weights as $q(t)=(1-t)p+tq^{\mathrm{ext}}$. The parameter $t$ controls the intervention strength: $t=0$ gives the reference weights, and we test values up to $0.64$. We choose a fixed imbalanced target weight vector $q^{\mathrm{ext}}$ that concentrates probability mass on the first few modes while retaining positive weight on every mode, where
\begin{equation}
\begin{aligned}
q^{\mathrm{ext}} = (&0.55,\,0.20,\,0.10,\,0.06,\\
                   &0.04,\,0.025,\,0.015,\,0.01).
\end{aligned}
\end{equation}
Interpolating toward this target varies coverage imbalance without changing mode locations or covariances. The candidate approximately commutes with $A$, and the measured displacement is predominantly spectral (\cref{fig:controlled-coverage-sweep}).

\paragraph{Mode merging:}
The parameter $\beta\in[0,1]$ controls how far the first two mode centers move toward their midpoint, keeping their weights and covariances fixed: $\beta=0$ leaves them unchanged, and $\beta=1$ merges them at the midpoint. The eight-dimensional RBF feature map remains fixed, with centers at the original
reference locations. Samples near the midpoint therefore activate both corresponding feature coordinates, giving a normalized response approximately equal to $(e_1+e_2)/\sqrt{2}$. Its outer product introduces off-diagonal entries between these coordinates, which approximately align with distinct reference
eigendirections at the chosen bandwidth. We observe that organizational energy dominates in the tested sweep
(\cref{fig:controlled-merge-sweep}).

\paragraph{Spurious mode:}
We add a Gaussian component at the ring center, with covariance $0.12^2I_2$ and mixture weight $f\in[0,0.38]$. The original eight components retain their relative weights and together receive total weight $1-f$. At the center, the normalized response is $\mathbf1/\sqrt8$; nearby samples also activate combinations of reference features. The measured displacement is predominantly organizational (\cref{fig:controlled-spurious-sweep}).

\begin{figure}[t]
    \centering

    \begin{subfigure}[b]{0.48\linewidth}
        \centering
        \includegraphics[width=\linewidth]
            {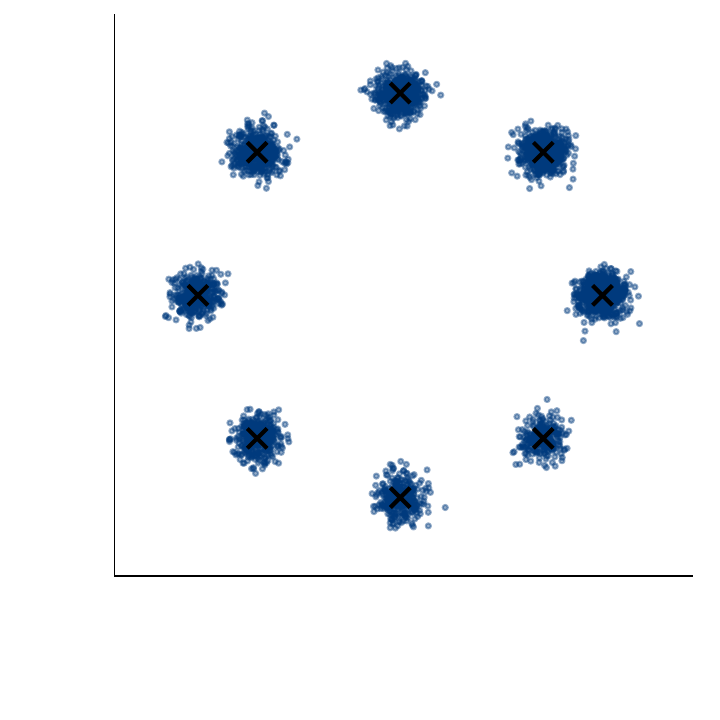}
        \caption{Reference}
        \label{fig:controlled-reference}
    \end{subfigure}\hfill
    \begin{subfigure}[b]{0.48\linewidth}
        \centering
        \includegraphics[width=\linewidth]
            {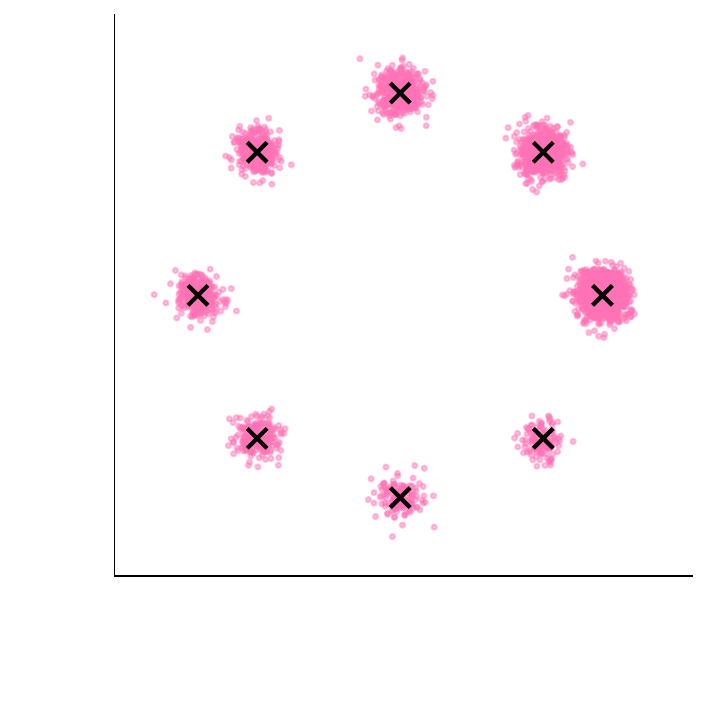}
        \caption{Coverage imbalance}
        \label{fig:controlled-coverage}
    \end{subfigure}

    \par\medskip

    \begin{subfigure}[b]{0.48\linewidth}
        \centering
        \includegraphics[width=\linewidth]
            {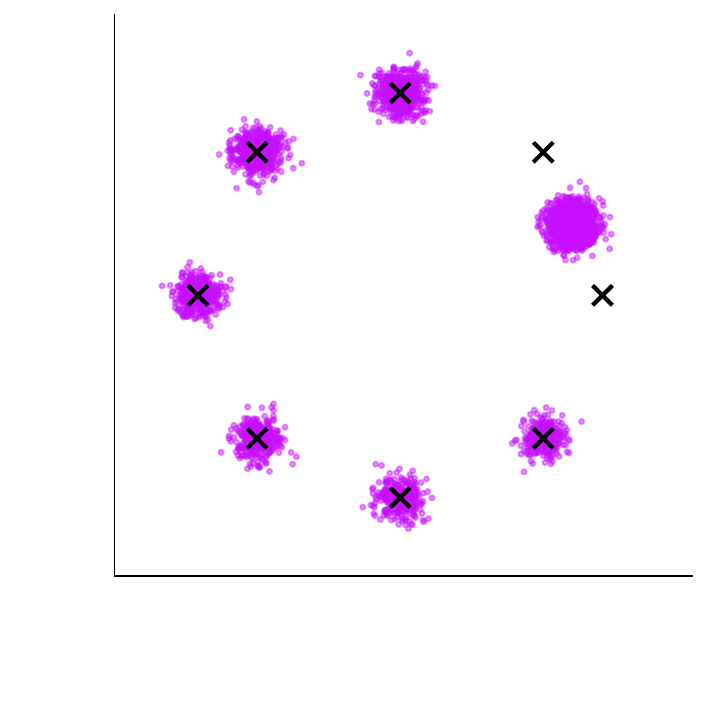}
        \caption{Mode merging}
        \label{fig:controlled-merging}
    \end{subfigure}\hfill
    \begin{subfigure}[b]{0.48\linewidth}
        \centering
        \includegraphics[width=\linewidth]
            {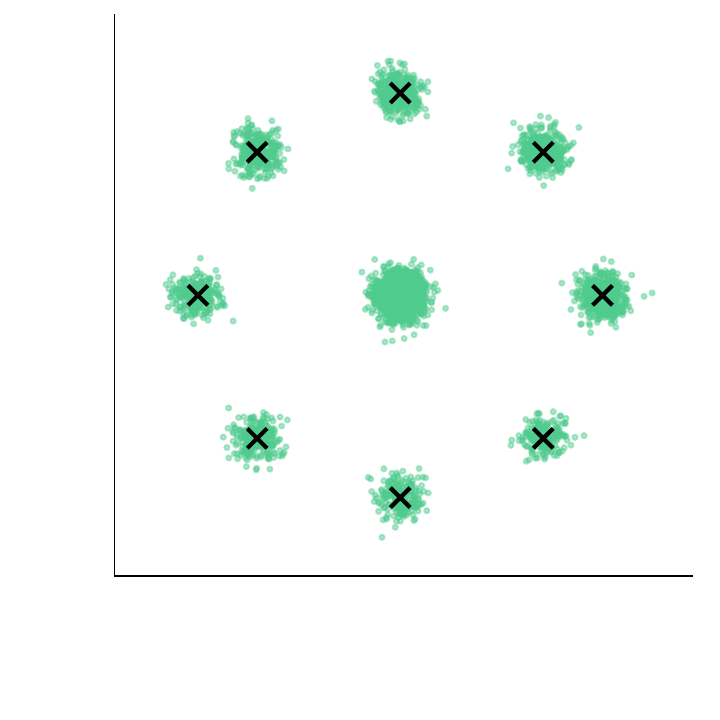}
        \caption{Spurious mode}
        \label{fig:controlled-spurious}
    \end{subfigure}

    \caption{
        \textbf{Controlled shifts with approximately matched total squared BW distance:}
        \textbf{(a)} Reference eight-mode Gaussian mixture.
        \textbf{(b)} Coverage imbalance changes mixture weights at fixed locations.
        \textbf{(c)} Mode merging combines neighboring modes.
        \textbf{(d)} Spurious-mode addition introduces a central mode.
        Black markers denote reference mode centers.
    }
    \label{fig:controlled-gmm}
\end{figure}

\subsubsection{Severity sweeps}
\label{app:controlled_severity_sweeps}

\label{app:controlled_sweeps}
\begin{figure}[t]
    \centering

    \begin{subfigure}[t]{\linewidth}
        \centering
        \includegraphics[width=\linewidth]
        {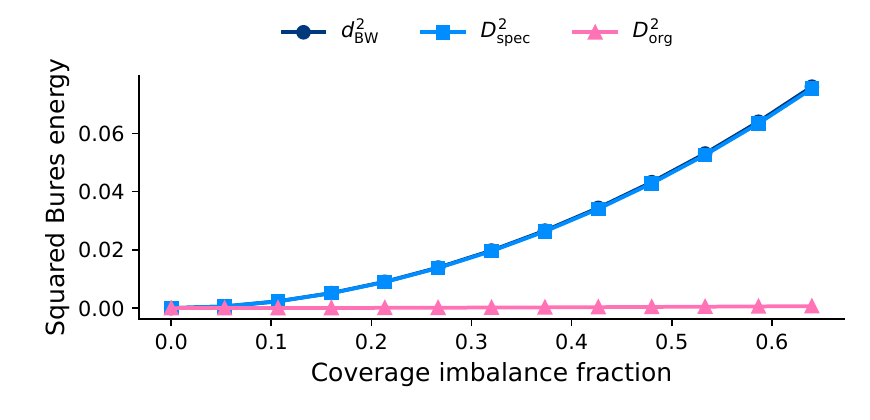}
        \caption{Coverage imbalance}
        \label{fig:controlled-coverage-sweep}
    \end{subfigure}

    \medskip
    \begin{subfigure}[t]{\linewidth}
        \centering
        \includegraphics[width=\linewidth]
        {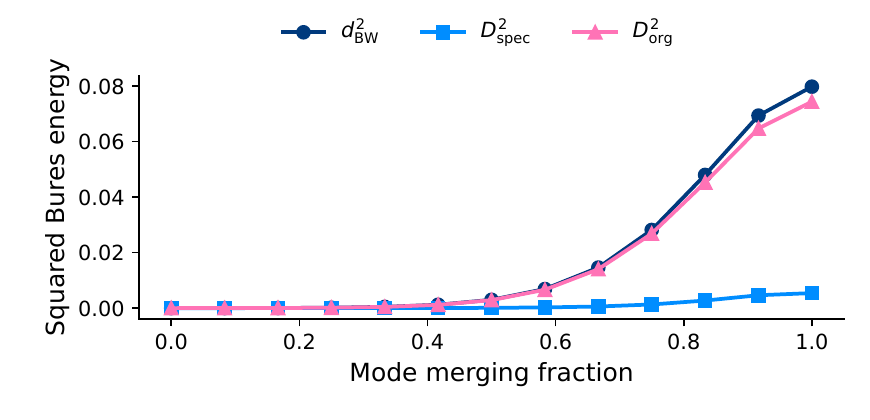}
        \caption{Mode merging}
        \label{fig:controlled-merge-sweep}
    \end{subfigure}

    \medskip
    \begin{subfigure}[t]{\linewidth}
        \centering
        \includegraphics[width=\linewidth]
        {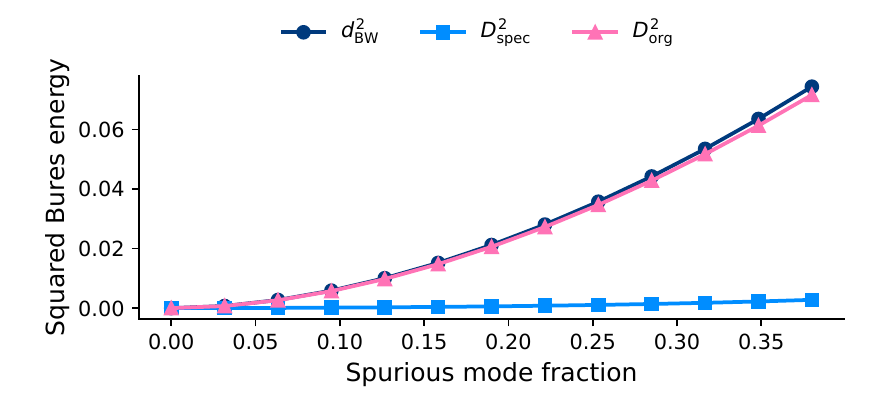}
        \caption{Spurious-mode addition}
        \label{fig:controlled-spurious-sweep}
    \end{subfigure}

    \caption{
        \textbf{Controlled perturbation sweeps:}
        At the default bandwidth and calibrated shrinkage, coverage imbalance gives predominantly spectral displacement, while mode merging and spurious-mode addition give predominantly organizational displacement. The plotted $D_{\mathrm{spec}}^2$ and $D_{\mathrm{org}}^2$ correspond to $E_{\mathrm{spec}}$ and $E_{\mathrm{org}}$ in the text.
    }
    \label{fig:controlled-sweeps}
\end{figure}

\cref{fig:controlled-sweeps} uses 13 equally spaced severity levels per intervention, with the same RBF feature map and shrinkage throughout. At nonzero severity, coverage imbalance remains predominantly spectral, while mode merging and spurious-mode addition remain predominantly organizational. Thus, the component split is observed across the severity sweeps, not only at the three endpoints reported in the main paper.

\subsubsection{Bandwidth and shrinkage sensitivity}
\label{app:rbf_sensitivity}

We assess sensitivity to the bandwidth of the normalized RBF feature map defined in \cref{app:feature_encoders}. Neighboring RBF centers are approximately $1.61$ units apart. With the default bandwidth $\sigma_{\mathrm{RBF}}=0.50$, samples near a reference center mainly activate its corresponding feature.

\paragraph{Bandwidth:}
We test bandwidths below and above the default to examine how the allocation changes as increasing bandwidth makes the RBF features overlap more strongly:
\begin{equation}
\begin{aligned}
\sigma_{\mathrm{RBF}}\in\{&0.20,\,0.30,\,0.40,\,0.50,\,0.65,\\
                        &0.85,\,1.10,\,1.40,\,1.80\}.
\end{aligned}
\end{equation}
We hold the mixture parameters, severity levels, calibration sample, and underlying Gaussian random draws fixed across bandwidths. At every tested bandwidth, calibration selected $\delta_{\min}=10^{-8}$, giving $\delta_{\mathrm{used}}=10^{-7}$.

At $\sigma_{\mathrm{RBF}}\in\{0.30,0.40,0.50,0.65\}$, coverage imbalance remains predominantly spectral and the other interventions remain predominantly organizational at every nonzero severity. The ordering changes as the features overlap more strongly: coverage imbalance is predominantly organizational at the tested bandwidths $0.85$ and above, and the spurious mode is predominantly spectral at $1.80$. At $0.20$, the near-zero spurious-mode result is affected by the implementation's $10^{-12}$ feature-normalization floor, which suppresses the central component's contribution. The controlled interpretation therefore holds over the tested bandwidths surrounding the default, but depends on the representation resolving the reference modes.

\paragraph{Shrinkage sensitivity:}
\label{app:delta_sensitivity}
\label{app:delta_sensitivity_controlled}
We reapply \cref{eq:isotropic_shrinkage} to the reference and candidate operators over a pre-specified grid. The primary range is
\begin{equation}
10^{-8}\le\delta\le5\times10^{-2},
\end{equation}
with $\delta\in\{0.1,0.2,0.5\}$ added to test stronger regularization. At each value, we recompute $d_{\mathrm{BW}}^2$, $E_{\mathrm{spec}}$, $E_{\mathrm{org}}$, and, for nonzero distance, the organizational share $\eta_{\mathrm{org}}=E_{\mathrm{org}}/d_{\mathrm{BW}}^2$.

Across the primary range, coverage imbalance remains predominantly spectral, with $\eta_{\mathrm{org}}=0.0090$--$0.0092$. Mode merging remains predominantly organizational, with $\eta_{\mathrm{org}}=0.932$--$0.938$, as does spurious-mode addition, with $\eta_{\mathrm{org}}=0.963$ throughout to three decimal places.
The same component dominates in each case at all three larger shrinkage values. Stronger shrinkage reduces the measured displacement as the operators approach $I_d/d$, but the qualitative separation among the three interventions remains.

\subsection{CIFAR-10 experiments}
\label{app:cifar_controlled}

\subsubsection{Data splits and paired interventions}
\label{app:cifar_data_splits}

We used 20 development, 10 calibration, and 30 evaluation blocks from the CIFAR-10 training split \cite{krizhevsky_learning_2009}. Each block contained
200 images, with 20 per class, and no image appeared in more than one block. Within each trial, the one-image-per-class reference was a subset of the five-image reference, which was a subset of the ten-image reference. Incoming images were disjoint from the reference and from the other incoming set. We used the same image selections for both encoders. These blocks were separate from the 4,000 test images used for shrinkage calibration and PCA. The prediction models
were fitted on development blocks and evaluated on evaluation blocks; the calibration blocks were not used in either step.

\paragraph{Paired interventions:}
Each reference contains five classes with one, five, or ten distinct images per class. For each reference we construct two incoming sets: ten fresh images from a single class already in the reference, or ten images from a single class absent from it. Let $P_{\mathrm{ref}}$ and $P_{\mathrm{in}}$ be the uniform empirical distributions over the reference and incoming images. We assign a fraction $\alpha$ of the endpoint's total weight to the incoming images:
\begin{equation}
P_\alpha=(1-\alpha)P_{\mathrm{ref}}+\alpha P_{\mathrm{in}}.
\end{equation}
The class is new or familiar relative to this reference, not relative to the encoder's training data.

\subsubsection{Fixed replacement and BW matching}
\label{app:cifar_bw_matching}
\paragraph{Fixed and matched comparisons:}
The fixed design uses $\alpha\in\{0.10,0.20,0.40\}$. The fixed-replacement curves (\cref{fig:cifar_clip_fixed}) use $\alpha=0.20$, while the pooled fixed-replacement prediction results use all three fractions. For each encoder, we selected one target squared BW distance from the attainable distance ranges on the development blocks. The choice used distance ranges, not spectral allocation or prediction performance. We then adjusted each endpoint's replacement fraction to reach that target. The targets for squared BW distance were $0.08105$ for DINOv2 and $0.02070$ for CLIP.

\paragraph{Matching checks:}
For each endpoint, we searched over $\alpha\in[0.01,0.80]$ using 20 bisection steps. An endpoint was accepted only if the achieved squared BW distance $D$ satisfied
\begin{equation}
\frac{|D-D_\star|}{D_\star}\le 0.01,
\end{equation}
where $D_\star$ is the encoder-specific target. A trial was retained only if both interventions at all three reference sizes passed. Each encoder retained all 150 evaluation trials across the 30 blocks, giving 900 matched endpoints: 150 for each intervention and reference size. No evaluation trial was excluded by the matching rule.

\subsubsection{Predictive models and statistical evaluation}
\label{app:cifar_prediction}
\paragraph{Prediction on held-out image blocks:}
Within each design, we compared ridge logistic regressions for the probability that the incoming class was absent. The baseline included squared BW distance, replacement-weighted mean nearest-neighbor ($k=1$) and centroid distances, reference and endpoint Vendi scores \cite{friedman_vendi_2023}, the distinct reference image count, Gaussian MMD$^2$ on normalized DINOv2 and CLIP features, and replacement-weighted official CMMD \citep{jayasumana_rethinking_2024}. We log-transformed the Vendi scores, squared BW distance, reference image count, and kernel discrepancies, then included each squared term and each pairwise product of the baseline predictors. The augmented model added spectral share and its interactions with log reference image count and log reference
Vendi score. For matched comparisons, every observation received the common target BW distance as its regression predictor. Small numerical differences between the achieved distances were therefore not used for prediction.

\paragraph{Kernel discrepancies and mixture weighting:}
We compute Gaussian-kernel MMD$^2$ on normalized features using the diagonal-including empirical estimator (i.e., retaining terms with $i=j$). For each encoder, we compute pairwise squared feature distances within the independent calibration reference subsets and pool these distances. The Gaussian bandwidth is set to the square root of their median \cite{gretton_kernel_2012} and remains fixed across comparisons. Separately, we compute CMMD using the official implementation of \cite{jayasumana_rethinking_2024}, with Gaussian bandwidth $\sigma=10$ and a scaling factor of $1000$. We use its CLIP image embeddings rather than the normalized features used for the BW operators.

For the endpoint mixture $P_\alpha=(1-\alpha)P_{\rm ref}+\alpha P_{\rm in}$, linearity of the kernel mean embedding gives
\begin{equation}
\begin{aligned}
\operatorname{MMD}^2(P_{\rm ref},P_\alpha)
&=\alpha^2\operatorname{MMD}^2(P_{\rm ref},P_{\rm in}).
\end{aligned}
\end{equation}
We apply this identity to both normalized-feature MMD$^2$ predictors and to CMMD, which is a scaled squared kernel discrepancy. Thus the CMMD predictor is $\alpha^2$ times the official score between the reference and incoming sets, accounting for the unchanged reference mass in the endpoint.

\paragraph{Fitting and evaluation:}
Predictors are standardized using only the fitting rows, separately within each cross-validation fold. Fitting weights give each block equal total weight. The ridge penalty applies to slope coefficients; the intercept is unpenalized. We select the penalty from $\{0.001,0.01,0.1,1\}$.

We selected the ridge penalty by five-fold cross-validation on development blocks, using mean block log loss, and evaluated both fitted models on the same 30 held-out evaluation blocks. For each block we subtracted the augmented model's mean Brier score \cite{brier_verification_1950} from the baseline's; positive values favor the added allocation features. Each block contributed equally to the reported mean. Within each encoder, we apply a Bonferroni correction \cite{dunn_multiple_1961} for the two pooled comparisons (fixed and matched), giving 97.5\% paired block-bootstrap intervals from 5,000 resamples of the 30 evaluation blocks. Each resample retains both models and all paired observations within the sampled blocks. The fitted models remain fixed during bootstrap resampling. 

The reference-size-specific intervals shown in \cref{fig:dino_brier_by_reference} and \cref{fig:clip_brier_by_reference} use 20,000 resamples at the same 97.5\% level and are descriptive. Specifically, the Bonferroni adjustment applies only to F/All and M/All, separately within each encoder; it does not provide simultaneous coverage across all subgroups or both encoders. We report these comparisons as exploratory.

\subsubsection{CLIP results}
\label{app:cifar_clip_results}

\begin{figure}[t]
    \centering
    \begin{subfigure}[t]{\linewidth}
        \centering
        \includegraphics[width=0.9\linewidth]
            {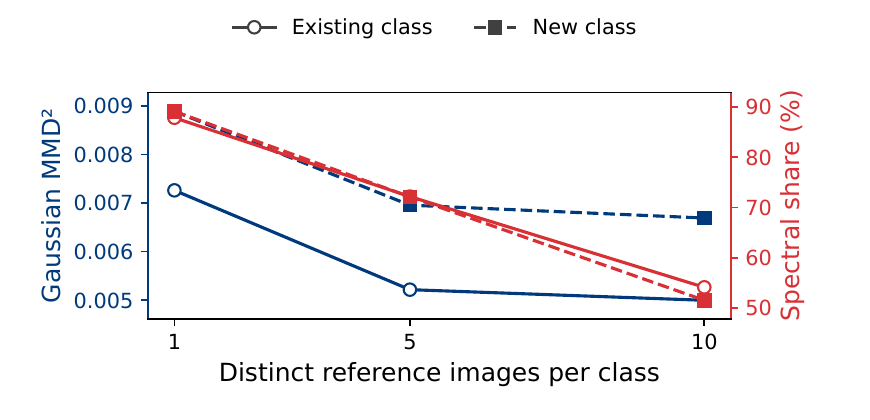}
        \caption{Fixed 20\% replacement}
        \label{fig:cifar_clip_fixed}
    \end{subfigure}\par\smallskip

    \begin{subfigure}[t]{\linewidth}
        \centering
        \includegraphics[width=0.9\linewidth]
            {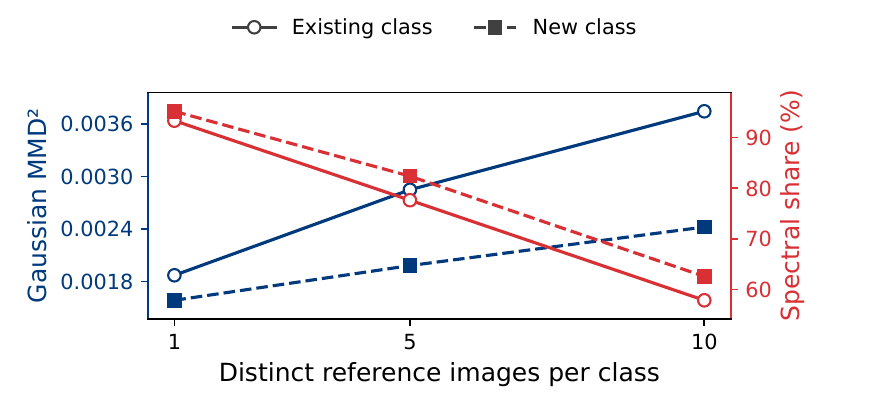}
        \caption{Matched BW distance}
        \label{fig:cifar_clip_matched}
    \end{subfigure}\par\smallskip

    \begin{subfigure}[t]{\linewidth}
        \centering
        \includegraphics[width=0.9\linewidth]
            {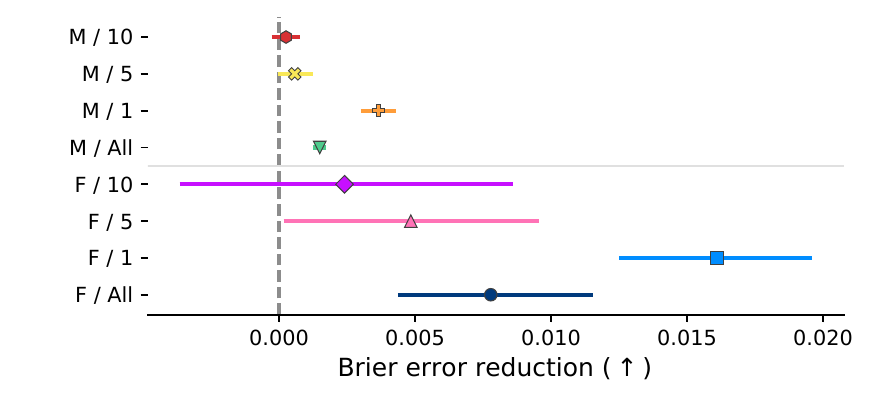}
        \caption{Predictive gain from spectral allocation}
        \label{fig:clip_brier_by_reference}
    \end{subfigure}

    \caption{\textbf{Reference diversity and predictive gain in CLIP:}
    (a--b) Mean spectral share (red, right axis) and Gaussian MMD$^2$ (blue, left axis) under fixed replacement and matched BW distance.
    (c) Reduction in held-out Brier error after adding spectral allocation to the baseline; positive values favor allocation. F denotes fixed replacement and M denotes matched BW. ``All'' pools reference sizes; 1, 5, and 10 denote distinct reference images per class. Bars show 97.5\% paired block-bootstrap intervals. For F/All and M/All, this level gives a Bonferroni correction for the two pooled comparisons within CLIP. Reference-specific comparisons are descriptive.}
    \label{fig:cifar_clip_fixed_matched}
\end{figure}

\paragraph{CIFAR-10: comparison across encoders:}
\cref{fig:cifar_clip_fixed_matched} shows the corresponding CLIP results: spectral share and Gaussian MMD$^2$ under fixed replacement and matched BW distance, together with the held-out Brier-score reduction from adding spectral allocation. As in DINOv2, mean spectral share decreases as reference size increases under both fixed replacement and BW matching. The pooled absolute reduction in Brier error is $0.0078$ under fixed replacement and $0.0015$ under BW matching, so the larger predictive gain again occurs under fixed replacement.

\subsubsection{Fixed-dimensional reference-size check}
\label{app:cifar_pca4}

Increasing reference size changes both the available information and the rank of the empirical operator. To examine the latter effect, we fitted a four-dimensional PCA projection separately for each encoder using 4,000 independent calibration images. The projection was fixed across reference sizes, projected features were normalized, and the original image selections and evaluation blocks were reused. For BW matching in the projected space, we selected a new target from the development blocks and adjusted each endpoint's replacement fraction using the procedure in \cref{app:cifar_bw_matching}. Every reference operator had rank four before shrinkage. Both encoders selected $\delta_{\min}=10^{-8}$ and used $\delta=10^{-7}$.

As in the full-dimensional experiment, mean spectral share was lower with ten reference images per class than with one, for both encoders and both interventions under fixed replacement and BW matching. All eight paired 95\% intervals for the ten-minus-one change were below zero (\cref{tab:cifar_pca4_effect}). These intervals are unadjusted, and the decrease was not always monotonic across the intermediate reference size.

\begin{table}[t]
\centering
\small
\setlength{\tabcolsep}{3pt}
\begin{tabular}{@{}lll r@{}}
\toprule
Encoder & Setting & Incoming & Change [95\% CI] \\
\midrule
DINOv2 & Fixed   & Familiar & $-4.46\,[-8.14,-0.80]$ \\
       & Fixed   & Absent   & $-9.56\,[-13.53,-5.61]$ \\
       & Matched & Familiar & $-5.49\,[-9.27,-1.71]$ \\
       & Matched & Absent   & $-11.87\,[-16.05,-7.47]$ \\
\midrule
CLIP   & Fixed   & Familiar & $-12.92\,[-15.95,-9.91]$ \\
       & Fixed   & Absent   & $-12.81\,[-16.25,-9.10]$ \\
       & Matched & Familiar & $-13.52\,[-16.58,-10.42]$ \\
       & Matched & Absent   & $-14.99\,[-18.92,-10.96]$ \\
\bottomrule
\end{tabular}
\caption{\textbf{Reference-size effect after four-dimensional PCA:}
Mean spectral share with ten reference images per class minus that with one, in percentage points. Negative values indicate lower share with ten images.
Fixed replacement uses $\alpha=0.2$. Brackets show unadjusted paired 95\% block-bootstrap intervals from 5,000 resamples of 30 evaluation blocks. All unregularized reference operators have rank four.}
\label{tab:cifar_pca4_effect}
\end{table}

All projected reference operators had rank four before shrinkage, which means that the decrease also occurs without a change in rank (\cref{tab:cifar_pca4_effect}). However, this comparison does not separate semantic coverage from estimation effects associated with more distinct reference images.

\subsubsection{Sample-size sensitivity}
\label{app:cifar_finite_stability}

We estimated operators using 5, 10, 25, 50, 100, 250, and 500 images per class, balanced across all ten classes with fixed shrinkage. At each size, we drew 100 subsets, sampling without replacement within each repetition, and measured the squared Frobenius difference from a fixed benchmark operator. The benchmark used 460 images per class (4,600 total), the largest balanced set available
from the cached features after reserving at least 500 disjoint images per class for subset sampling.

From 5 to 500 images per class, the mean squared Frobenius difference decreased from $0.0180$ to $0.000368$ for DINOv2 and from $0.00577$ to $0.000136$ for CLIP. This indicates closer agreement with the separate empirical benchmark as sample size increases. Since the benchmark is itself a finite-sample estimate, these differences do not measure population-estimation error or establish stability of the decomposition.
\section{Diffusion experiments}
\label{app:cfg_additional}

\subsection{SDXL experimental protocol}
\label{app:sdxl_protocol}
The prompts are drawn from the GenEval two-object prompt set \cite{ghosh_geneval_2023}. We define a prompt family by its unordered pair of requested categories; the selected 40 prompts belong to 40 distinct families.

We use SDXL Base 1.0 \cite{podell_sdxl_2024} with 40 two-object prompts and 32 seeds per prompt, generating at $1024\times1024$ resolution. Sampling uses 32 deterministic Heun updates with an EDM noise schedule \cite{karras_elucidating_2022} ($\sigma_{\min}=0.0292$, $\sigma_{\max}=14.6146$, $\rho=3$), following the SDXL sampling setup used in \cite{kynkaanniemi_applying_2024}. The three branches share the same initial latent and null-conditioned trajectory through state $t=10$. Null remains unconditioned; Prompt introduces text conditioning at update 11 with guidance scale 1 for the remaining updates. CFG introduces the same prompt and uses guidance scale 16 during updates 11--26, then scale 1 thereafter.

To obtain comparable images at intermediate states, we apply the same null-conditioned clean-image readout to every branch: one denoiser evaluation at the current noise level predicts a clean latent, which is decoded by the VAE. At the final state, we decode the latent directly. Intermediate readouts are therefore clean-image predictions from the current state, not trajectories continued to the final state. Decoded images are resized to $512\times512$ using Pillow's Lanczos resampling and saved as PNGs before feature extraction and detection.

We encode the same saved images using DINOv2 ViT-S/14 and CLIP ViT-L/14@336, yielding 384- and 768-dimensional features, respectively. Feature preprocessing and operator construction follow \cref{app:feature_encoders}, with shrinkage calibrated as described in \cref{app:calibration_sets}. Each trajectory is compared with its prompt-specific state-0 reference.

Detection uses RT-DETR \cite{zhao_detrs_2024} with the \texttt{PekingU/rtdetr\_r50vd} checkpoint and a confidence threshold of $0.5$. Each requested category is scored separately; an image receives a score of $0$, $1/2$, or $1$ according to whether neither, one, or both requested categories are detected. Reported detection rates average these scores over prompts and seeds.

\subsection{Additional trajectories and CLIP results}
\label{app:sdxl_additional_trajectories}

\begin{figure}[t]
    \centering
    \begin{subfigure}[t]{\linewidth}
        \centering
        \includegraphics[width=\linewidth]{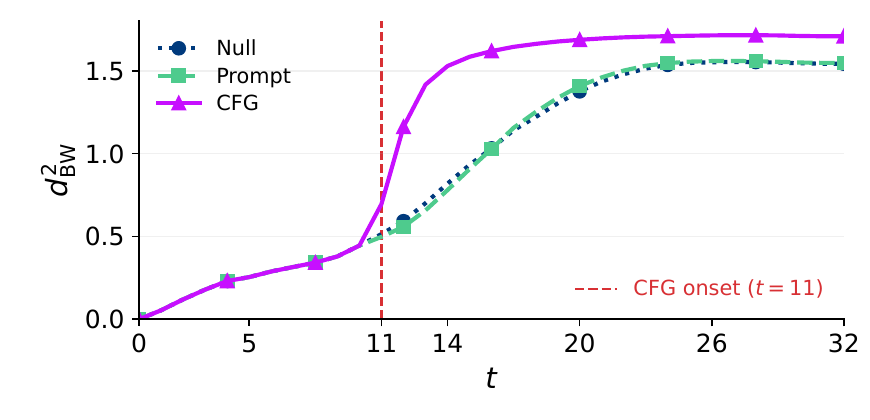}
        \caption{DINOv2}
        \label{fig:cfg_switch_bw}
    \end{subfigure}\par\smallskip
    \begin{subfigure}[t]{\linewidth}
        \centering
        \includegraphics[width=\linewidth]
            {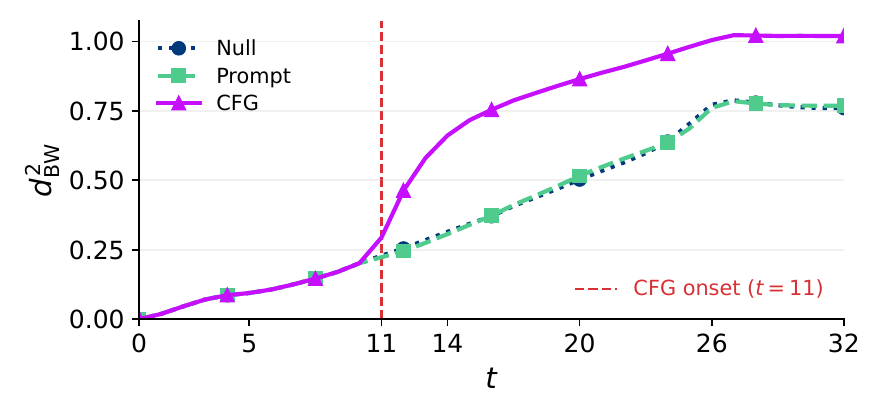}
        \caption{CLIP}
        \label{fig:app_cfg_clip_bw}
    \end{subfigure}
    \caption{\textbf{SDXL displacement from the state-0 reference:}
    Curves show mean squared BW distance for the Null, Prompt, and CFG branches over 40 prompts with 32 seeds per prompt, using the same images in both encoder analyses. The branches share the same latent through $t=10$; the red dashed line marks the first guided update at $t=11$. The panels use different vertical scales.}
    \label{fig:app_cfg_bw_encoders}
\end{figure}

\cref{fig:app_cfg_bw_encoders} shows the squared BW distance from the state-0 reference in both feature spaces. The distance changes over the trajectory, so its contribution to the spectral-energy floor also changes. We therefore subtract $D^2/4$ from each branch's spectral energy before comparing CFG with Prompt, where $D=d_{\mathrm{BW}}^2$.

\begin{figure}[t]
    \centering
    \begin{subfigure}[t]{\linewidth}
        \centering
        \includegraphics[width=0.9\linewidth]
            {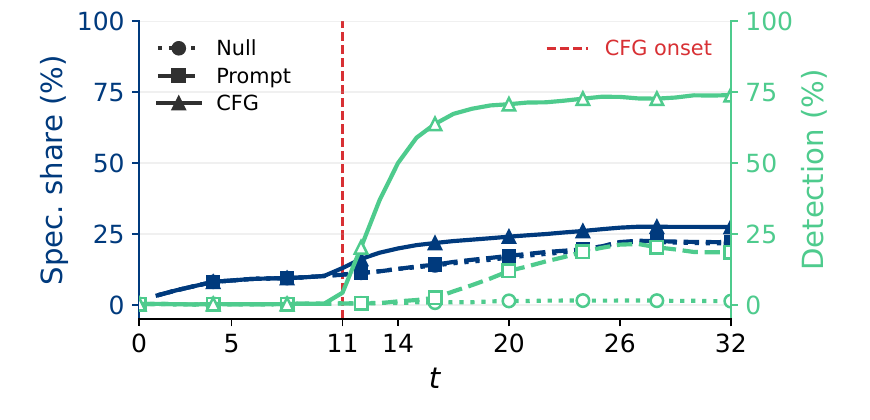}
        \caption{Spectral share and requested-category detection}
        \label{fig:cfg_clip_share_detection}
    \end{subfigure}\par\smallskip
    \begin{subfigure}[t]{\linewidth}
        \centering
        \includegraphics[width=0.9\linewidth]
            {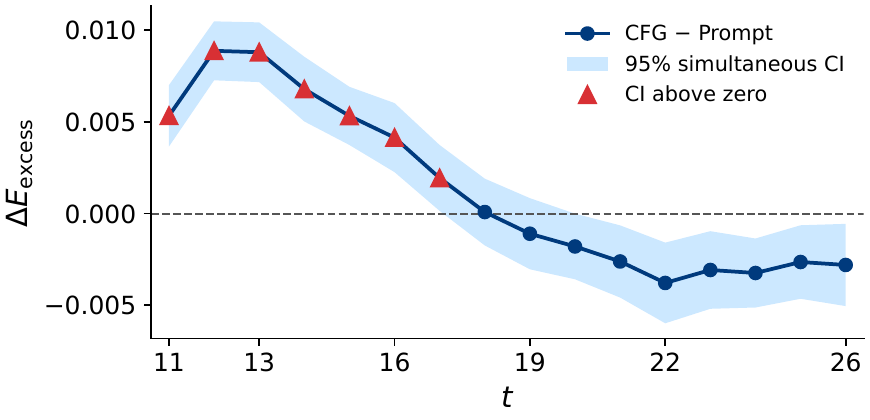}
        \caption{CFG-minus-Prompt difference in excess spectral energy}
        \label{fig:cfg_clip_excess_band}
    \end{subfigure}
    \caption{\textbf{Guidance onset in CLIP feature space:}
    (a) Filled blue markers show spectral share relative to the fixed state-0 reference; hollow green markers show the fraction of requested categories detected in the same images. Branches are identified by marker shapes and line styles. The red dashed line marks the first guided update ($t=11$).
    (b) $\Delta E_{\mathrm{excess}}$ is the mean CFG-minus-Prompt difference in $E_{\mathrm{spec}}-D^2/4$, where $D$ is squared BW distance. Shading shows the 95\% simultaneous prompt-family bootstrap band for the mean CFG-minus-Prompt difference over all 16 guided updates. Red triangles indicate updates where the lower bound exceeds zero. Both panels use 40 prompts with 32 seeds each.}
    \label{fig:cfg_clip_trajectory}
\end{figure}

The CLIP counterpart in \cref{fig:cfg_clip_excess_band} also shows an early positive excess-energy response followed by a decline. Its 95\% simultaneous band is above zero at updates 11--17 and below zero at updates 20--26. The mean peaks at $t=12$ at $0.00887$. Thus, CLIP supports the early-response pattern observed in DINOv2, although the peak timing and magnitude differ. \cref{fig:cfg_guided_excess_band} reports the DINOv2 guided-interval band.

\paragraph{Uncertainty across prompt families:}
For the excess-energy contrasts in \cref{fig:cfg_guided_excess_band} and \cref{fig:cfg_clip_excess_band}, we first compute the paired CFG-minus-Prompt excess-energy contrast for each prompt at every update. We construct a 95\% simultaneous confidence band using a studentized max-\(t\) bootstrap \cite{romano_stepwise_2005}. For each of 20,000 bootstrap resamples, we draw 40 prompt families with replacement, keeping all states and paired branches together. At each update, we calculate the mean of their paired CFG-minus-Prompt contrasts and its standard error. We subtract the original mean curve from the resampled mean curve, divide by the resampled standard error at each update, and retain the largest absolute value across updates 11--26. The 95th percentile of these values determines the simultaneous band: the original mean curve plus or minus this multiplier times the original standard error at each update.

The same procedure is used for the centering check and FLUX trajectories. These intervals condition on the saved seed samples, estimated operators, and fixed shrinkage: seeds are not resampled and shrinkage is not recalibrated. \cref{tab:sdxl_shrinkage} instead reports pointwise percentile bootstrap intervals, without simultaneous coverage across updates or shrinkage values.

\subsection{Guidance-schedule comparisons}
\label{app:sdxl_schedule_comparisons}

We compare five schedules using the same 40 prompts and 32 seeds per prompt. All share the null-conditioned prefix during updates $1 \leq t \leq 10$ and use text conditioning thereafter:
\begin{itemize}
    \item \textbf{Early-only (E):} guidance scale 16 during
    updates 11--15.
    \item \textbf{Late-only (L):} guidance scale 16 during
    updates 22--26.
    \item \textbf{Full-window (F):} guidance scale 16 during
    updates 11--26.
    \item \textbf{Delayed (D):} guidance scale 16 during
    updates 16--26.
    \item \textbf{Prompt-only (P):} guidance scale 1
    throughout the post-prefix updates.
\end{itemize}
For E, L, F, and D, guidance scale is 1 at all post-prefix updates outside the specified window.

\cref{fig:cfg_schedule_detection} compares five guidance schedules using the same 40 prompts and 32 seeds per prompt. Using the detection scores defined in \cref{app:sdxl_protocol}, we average over seeds within each prompt and compute paired schedule differences. The 95\% percentile intervals use 10,000 prompt-family resamples, retaining all seeds and schedule pairings within each family. These are individual contrast intervals, without a joint correction across the five schedule contrasts. For the five-percentage-point noninferiority check, we compare the fifth percentile of the bootstrap E--F difference with $-5$ percentage points. 

\subsection{SDXL robustness checks}
\label{app:sdxl_stability}

We used saved DINOv2 features from the 32-seed experiment to test eigenvalue grouping and basis rotations on three prompts, and shrinkage sensitivity on all 40 prompts. The centering and sample-size checks use all 40 prompts with both DINOv2 and CLIP.

\subsubsection{Eigenvalue grouping and basis invariance}
\label{app:sdxl_eigenspace_checks}
We repeated the decomposition with grouping tolerances of $0.1$, $1$, and $10$ times the default. For all three prompts, the reference eigenvalue groups were identical at all three tolerances. Spectral share was unchanged across all three branches at states 10--26 and 32. This tests the numerical grouping choice at the observed reference operators.

We also tested whether the numerical result depended on the basis chosen within a repeated eigenspace. For each of the three prompts, we applied five random rotations within the 352-dimensional repeated reference eigenspace and recomputed the component energies for the CFG branch at $t=11$. The largest absolute change was $5.6\times10^{-16}$, consistent with roundoff and the basis invariance in \cref{prop:organizational-faithfulness}. For the three audited prompts, the results were insensitive to the tested grouping tolerances and basis rotations. However, these checks do not establish stability when eigenvalues approach
a grouping boundary.

\subsubsection{Shrinkage sensitivity}
\label{app:sdxl_shrinkage_sensitivity}
We recomputed the DINOv2 trajectories at five shrinkage values from $10^{-5}$ to $10^{-3}$. All reference and endpoint operators met the conditioning threshold. At the first guided update ($t=11$), the mean CFG-minus-Prompt difference in excess spectral energy was positive at every shrinkage value. Its pointwise 95\% interval included zero at the two smallest values. From $t=12$ through $t=15$, the intervals lay entirely above zero for all five values, although the magnitude of the difference varied with shrinkage (\cref{tab:sdxl_shrinkage}). The early positive response persists over this range, while its size and the first update with a pointwise interval above zero depend on shrinkage.

\begin{table*}[t]
    \centering
    \small
    \setlength{\tabcolsep}{4pt}
    \begin{tabular}{@{}lrrrrr@{}}
        \toprule
        $\delta$ & $t=11$ & $t=12$ & $t=13$ & $t=14$ & $t=15$ \\
        \midrule
        $10^{-5}$
        & $0.17\,[-1.12, 1.54]$
        & $7.52\,[5.89, 9.23]$
        & $10.18\,[8.59, 11.99]$
        & $10.51\,[8.45, 12.67]$
        & $8.57\,[6.31, 10.92]$ \\

        $3\times10^{-5}$
        & $0.78\,[-0.63, 2.24]$
        & $9.32\,[7.62, 11.11]$
        & $11.68\,[9.98, 13.56]$
        & $12.01\,[9.82, 14.25]$
        & $9.40\,[6.97, 11.82]$ \\

        $10^{-4}$
        & $2.05\,[0.53, 3.68]$
        & $12.70\,[10.81, 14.68]$
        & $14.74\,[12.86, 16.84]$
        & $14.82\,[12.38, 17.22]$
        & $11.18\,[8.60, 13.67]$ \\

        $3\times10^{-4}$
        & $4.39\,[2.51, 6.44]$
        & $18.28\,[16.00, 20.65]$
        & $20.21\,[17.97, 22.68]$
        & $19.27\,[16.49, 21.93]$
        & $14.45\,[11.60, 17.23]$ \\

        $10^{-3}$
        & $8.84\,[6.20, 11.75]$
        & $29.36\,[26.44, 32.45]$
        & $31.66\,[28.81, 34.70]$
        & $28.32\,[24.82, 31.59]$
        & $21.49\,[17.85, 25.03]$ \\
        \bottomrule
    \end{tabular}
    \caption{\textbf{Sensitivity of excess spectral energy to shrinkage:}
    Entries give the mean CFG-minus-Prompt difference in DINOv2 excess spectral energy over 40 prompts with 32 seeds per prompt. Estimates and interval endpoints are multiplied by $10^3$. Brackets give pointwise 95\% percentile intervals from 5,000 prompt-family bootstrap resamples, preserving the pairing of CFG and Prompt within each prompt. These intervals are unadjusted for comparisons across updates or shrinkage values. The primary DINOv2 analysis uses $\delta=10^{-4}$.}
    \label{tab:sdxl_shrinkage}
\end{table*}

\subsubsection{Centered and uncentered operators}
\label{app:sdxl_centering}

\label{app:centered_covariance}
The main analysis uses uncentered second moments. Centering the features within each collection removes the contribution of its mean: for feature mean $\mu$, the centered operator before trace normalization is $M-\mu\mu^\top$.

We compare centered and uncentered operators using the same 40 prompts and 32 seeds per prompt, processing each reference and endpoint consistently. Centering is followed by trace normalization and the same fixed shrinkage used for the uncentered analysis.

\begin{figure}[t]
    \centering
    \begin{subfigure}[t]{\linewidth}
        \centering
        \includegraphics[width=\linewidth]
            {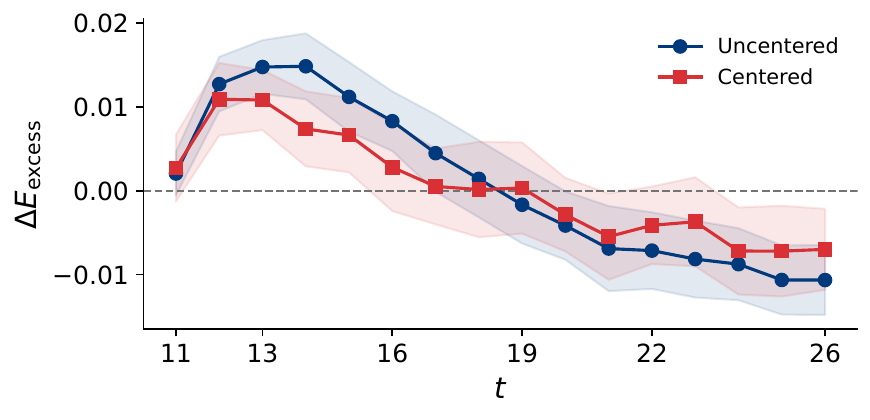}
        \caption{DINOv2}
    \end{subfigure}\par\smallskip

    \begin{subfigure}[t]{\linewidth}
        \centering
        \includegraphics[width=\linewidth]
            {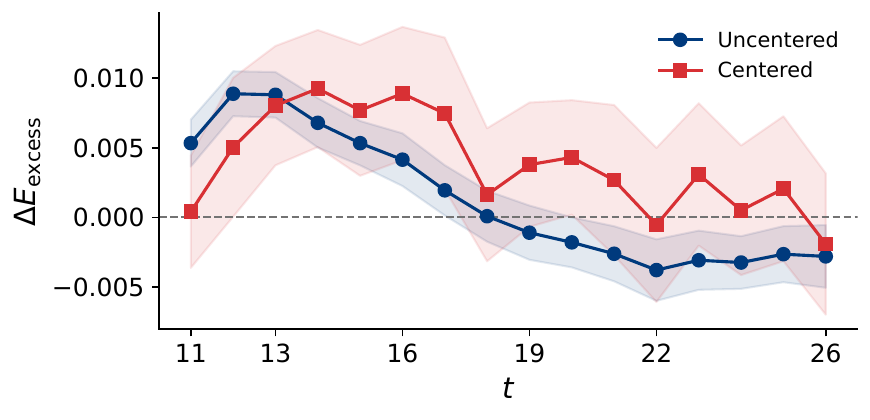}
        \caption{CLIP}
    \end{subfigure}

    \caption{\textbf{Effect of centering on spectral allocation:}
    Curves show the mean CFG-minus-Prompt excess spectral energy, $\Delta E_{\mathrm{excess}}$, for uncentered and centered operators using the same 40 prompts
    and 32 seeds per prompt. Shading shows 95\% simultaneous bands over updates 11--26, separately for each encoder and operator construction, not jointly across all four curves. The panels use different vertical scales.}
    \label{fig:sdxl_centered_uncentered}
\end{figure}

\cref{fig:sdxl_centered_uncentered} compares the two operator constructions for both encoders. For DINOv2, centering preserves the early positive CFG-minus-Prompt excess-energy response but reduces its peak. For CLIP, centering changes the trajectory more substantially, particularly later in the guided interval. The allocation therefore depends on whether mean-feature information is retained.

\subsubsection{Operator sample-size sensitivity}
\label{app:sdxl_finite_stability}
For $n$ independent, unit-normalized features, \cref{lem:operator_mse} in \cref{app:operator_estimation_error} bounds the expected squared Frobenius error of
the empirical second-moment operator by $1/n$. Applying the same fixed shrinkage to the empirical and population operators reduces the bound to $(1-\delta)^2/n$
by \cref{lem:operator_mse}. This controls operator estimation error but does not by itself establish stability of the estimated eigenspaces or energy allocation.

We measured how the operator estimates varied with sample size. For each prompt, we repeated the subset experiment 100 times. In each repetition, we selected nested subsets of 4, 8, 12, 16, 24, and 32 seeds without replacement, so each smaller subset was contained in the next larger one. We held shrinkage fixed and measured the Frobenius distance from each subset estimate to the corresponding 32-seed operator. For both encoders, larger subsets gave closer agreement with this benchmark. For DINOv2, the mean distance from 4 to 12 seeds fell from $0.312$ to $0.153$ for state-0 references and from $0.421$ to $0.206$ for comparison
operators, pooled over Prompt and CFG at states 11, 12, 13, 15, 26, and 32 across all 40 prompts. These distances measure sensitivity to using fewer of the
observed seeds. They vanish at $n=32$ since each estimate then equals its benchmark. The test therefore does not measure population-estimation error or establish stability of the component energies.

\subsection{Cross-model check with FLUX}
\label{app:flux_guidance}

\subsubsection{Experimental protocol}
\label{app:flux_protocol}

We use FLUX.2 [klein] Base 4B \cite{black_forest_labs_flux2_2026} with the same 40 two-object prompts and 32 seeds per prompt as SDXL. This checks
whether the findings extend to another generator on the same prompt set. The checkpoint is \texttt{black-forest-labs/}\allowbreak \texttt{FLUX.2-klein-base-4B}. Sampling uses 32 updates with the model's native flow-matching Euler sampler and noise schedule, rather than the Heun sampler used for SDXL.

Branch definitions, the shared prefix through state \(t=10\), guidance windows, and pairing by prompt and initial latent follow the SDXL protocol in \cref{app:sdxl_protocol,app:sdxl_schedule_comparisons}.

For intermediate states, the empty-prompt velocity prediction gives the clean-latent readout
\begin{equation}
\widehat{x}_0(t)
= x_t-\sigma_t v_{\varnothing}(x_t,\sigma_t).
\end{equation}
We decode this estimate with the VAE, while we decode the accepted latent directly at the final state.

Feature extraction, operator construction, and comparisons with the state-0 reference follow \cref{app:sdxl_protocol}. Independent FLUX shrinkage calibration follows \cref{app:calibration_sets}, giving \(\delta=10^{-4}\) for DINOv2 and \(\delta=3\times10^{-4}\) for CLIP (\cref{tab:calibrated_shrinkage}). Detection scoring follows \cref{app:sdxl_protocol}. Trajectory bands and final-detection comparisons use the procedures in \cref{app:sdxl_additional_trajectories,app:sdxl_schedule_comparisons}, respectively, including the same noninferiority criterion.

\subsubsection{Trajectory and guidance-schedule results}
\label{app:flux_results}

\begin{figure}[t]
    \centering
    \begin{subfigure}[t]{\linewidth}
        \centering
        \includegraphics[width=\linewidth]
            {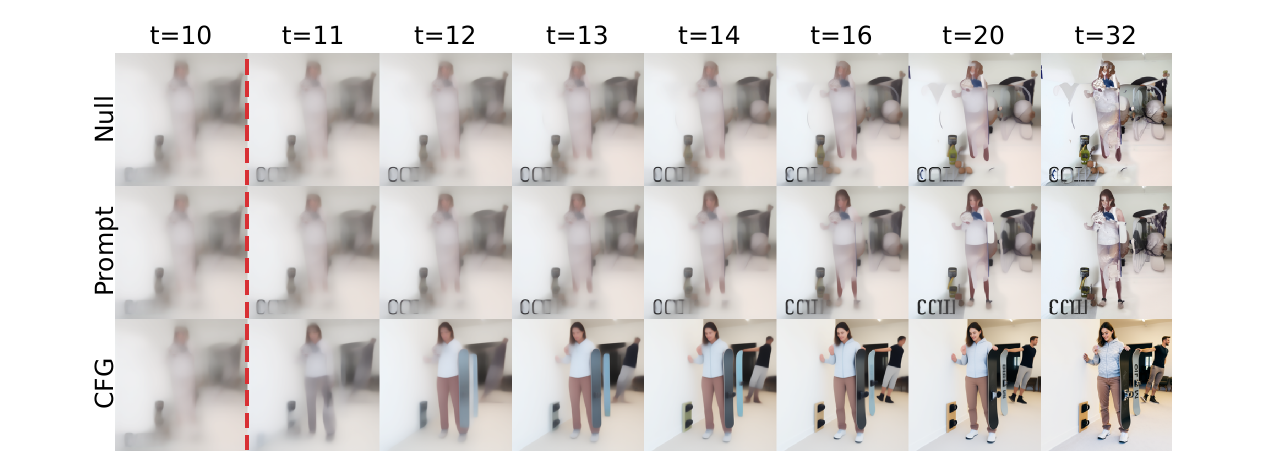}
        \caption{Shared image readouts}
        \label{fig:flux_images}
    \end{subfigure}\par\smallskip

    \begin{subfigure}[t]{\linewidth}
        \centering
        \includegraphics[width=\linewidth]
            {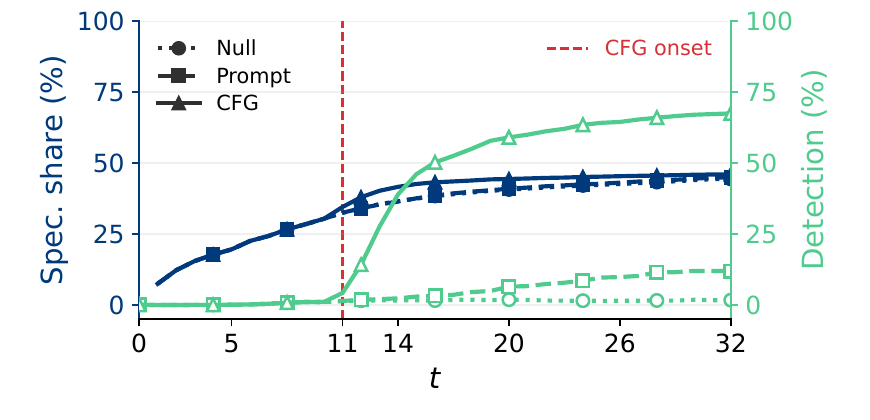}
        \caption{DINOv2}
        \label{fig:flux_dino_trajectory}
    \end{subfigure}\par\smallskip

    \begin{subfigure}[t]{\linewidth}
        \centering
        \includegraphics[width=\linewidth]
            {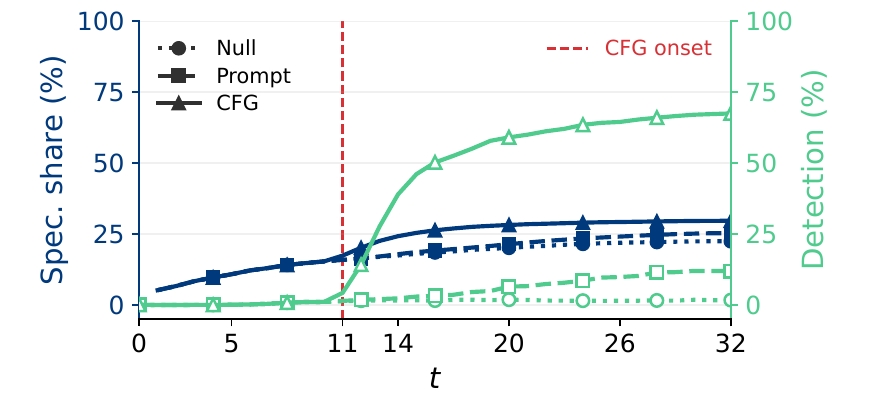}
        \caption{CLIP}
        \label{fig:flux_clip_trajectory}
    \end{subfigure}

    \caption{\textbf{FLUX trajectories and spectral allocation:}
    (a) Image readouts for ``a photo of a person and a snowboard,'' shared by both encoder analyses.
    (b--c) Filled blue markers show spectral share relative to the fixed state-0 reference (left axis); hollow green markers show requested-category detection (right axis). Marker shapes and line styles identify the branches. The red dashed line marks guidance onset at $t=11$. Curves summarize 40 prompts with 32 seeds each; detection values are identical across the two encoders.}
    \label{fig:flux_trajectories}
\end{figure}

\cref{fig:flux_trajectories} shows the FLUX image readouts, spectral shares, and requested-category detection rates. Under CFG, detection increases from $4.3\%$ at $t=11$ to $38.9\%$ at $t=14$, reaching $67.5\%$ at the final state.

\begin{figure}[t]
    \centering
    \begin{subfigure}[t]{\linewidth}
        \centering
        \includegraphics[width=\linewidth]
            {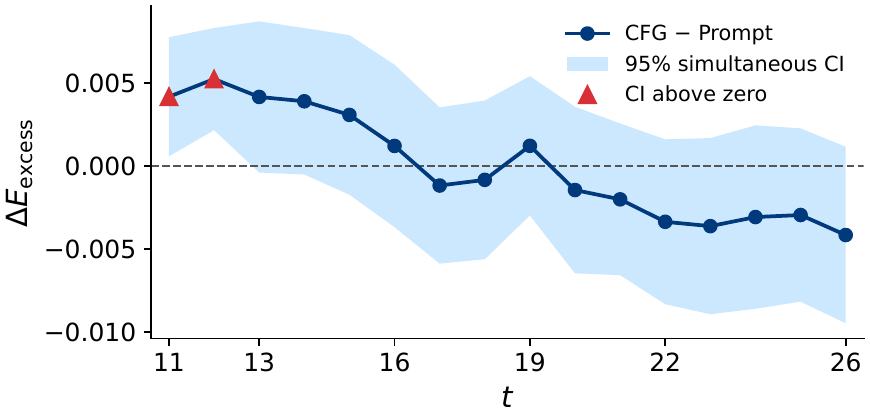}
        \caption{DINOv2 excess-energy contrast}
    \end{subfigure}\par\smallskip

    \begin{subfigure}[t]{\linewidth}
        \centering
        \includegraphics[width=\linewidth]
            {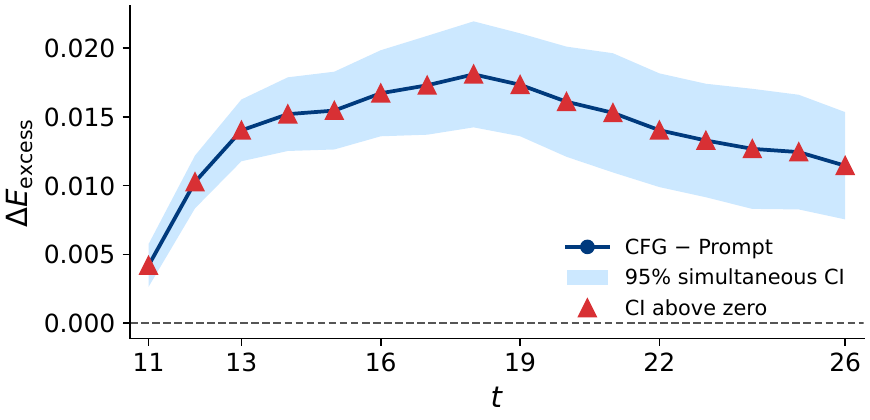}
        \caption{CLIP excess-energy contrast}
    \end{subfigure}\par\smallskip

    \begin{subfigure}[t]{\linewidth}
        \centering
        \includegraphics[width=\linewidth]
            {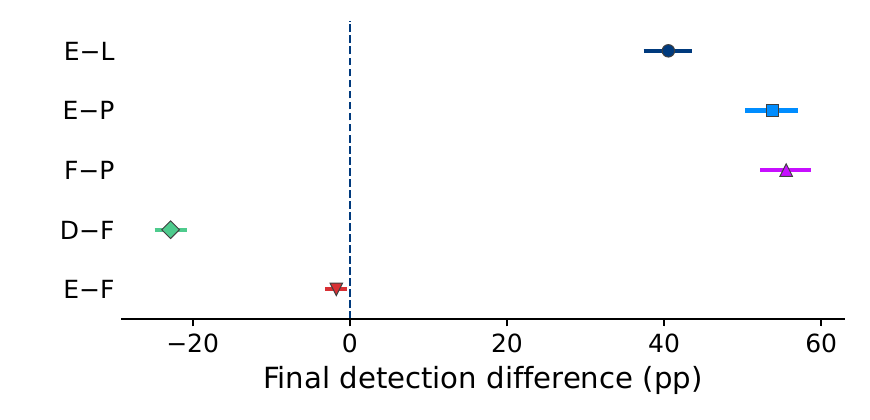}
        \caption{Paired differences in final detection}
    \end{subfigure}

    \caption{\textbf{FLUX excess spectral energy and guidance timing:}
    (a--b) $\Delta E_{\mathrm{excess}}$ denotes the mean CFG-minus-Prompt excess spectral energy. Shading shows a 95\% simultaneous band over updates
    11--26, separately for each encoder; red triangles indicate where its lower bound exceeds zero.
    (c) Final requested-category detection under the first schedule minus that under the second, in percentage points. E guides at updates 11--15; L at 22--26;
    F at 11--26; and D at 16--26. P uses prompt conditioning alone. Bars show paired 95\% prompt-family bootstrap intervals. All panels use 40 prompts with 32 seeds per prompt. The detection comparison is shared by both encoders.}
    \label{fig:flux_guidance_comparison}
\end{figure}

The excess-energy contrast depends on the encoder (\cref{fig:flux_guidance_comparison}). For DINOv2, the mean CFG-minus-Prompt contrast peaks at $t=12$ at $0.00525$, and the 95\% simultaneous band is above zero at updates 11--12. The mean becomes negative later, but the band includes zero at every update from 13 through 26. For CLIP, the contrast remains positive throughout updates 11--26, with the simultaneous band above zero over the entire interval. Its mean peaks later, at $t=18$, at $0.01809$. Thus, both encoders show an early positive response, but its timing and persistence differ.

The schedule comparison supports the early-over-late advantage observed in SDXL. Final requested-category detection is $65.7\%$ under E and $25.2\%$ under L,
a paired gain of $40.5$ percentage points (95\% prompt-family bootstrap CI $[37.4,\,43.5]$). Full guidance achieves $67.5\%$, compared with $44.6\%$ under delayed guidance and $11.9\%$ under prompt conditioning alone. Early guidance is $1.7$ percentage points below full guidance (E--F: 95\% CI $[-3.1,\,-0.4]$).
Thus, FLUX meets the five-percentage-point noninferiority criterion: the one-sided 95\% lower bound for E--F is $-2.9$ percentage points.

These comparisons support the benefit of early guidance under the tested protocol, but do not establish that the diagnostic selects better schedules than alternative selection rules.

\section{Proofs and auxiliary theoretical results}
\label{app:proofs}

We first establish regularization and finite-sample error bounds for the empirical operators, then collect the geometric identities used in the proofs. We then prove \cref{thm:finite-pair-decomposition,thm:spectral-trace-one} and \cref{prop:organizational-faithfulness} in
\cref{app:proof_prop1,app:trace_one_spectral_energy,app:proof_prop2}, respectively. The final subsection checks that the decomposition recovers its local limit.
All tangent spaces below are those of the ambient manifold $\mathbb S_{++}^d$, even when the endpoint operators have trace one.

\subsection{Regularized operators and BW norm bounds}
\label{app:regularized_operator_bounds}

\begin{lemma}[Spectral and metric bounds under isotropic shrinkage]
\label{lem:regularized_operator_bounds}
Fix $0<\delta<1$ and define
\begin{equation}
\mathcal D_{d,\delta}
:=
\left\{
A\in\operatorname{Sym}(d):
\operatorname{tr}(A)=1,\;
A\succeq\frac{\delta}{d}I_d
\right\}.
\end{equation}

For every $\rho\succeq0$ with $\operatorname{tr}(\rho)=1$, the regularized operator
\begin{equation}
\rho_\delta=(1-\delta)\rho+\frac{\delta}{d}I_d
\end{equation}
belongs to $\mathcal D_{d,\delta}\subset\mathbb S_{++}^d$.

Writing
\begin{equation}
m_\delta:=\frac{\delta}{d},
\qquad
M_\delta:=1-(d-1)\frac{\delta}{d},
\end{equation}
every $A\in\mathcal D_{d,\delta}$ satisfies
\begin{equation}
\begin{aligned}
m_\delta
&\le \lambda_{\min}(A) \\
&\le \lambda_{\max}(A) \\
&\le M_\delta.
\end{aligned}
\end{equation}

Consequently, for every $H\in\operatorname{Sym}(d)$,
\begin{equation}
\frac{1}{4M_\delta}\|H\|_F^2
\le \|H\|_A^2 \le \frac{1}{4m_\delta}\|H\|_F^2.
\end{equation}

\end{lemma}

\begin{proof}
Since $\rho\succeq0$ and $\operatorname{tr}(\rho)=1$,
\begin{equation}
\rho_\delta\succeq\frac{\delta}{d}I_d\succ0,
\qquad
\operatorname{tr}(\rho_\delta)=1.
\end{equation}

Thus $\rho_\delta\in\mathcal D_{d,\delta}$. For any $A\in\mathcal D_{d,\delta}$, each eigenvalue is at least $m_\delta$. Since their sum is one,
\begin{equation}
\begin{aligned}
\lambda_{\max}(A)
&\le 1-(d-1)m_\delta\\
&=M_\delta.
\end{aligned}
\end{equation}

Let $A=U\operatorname{diag}(\lambda_1,\ldots,\lambda_d)U^\top$ and $\widetilde H=U^\top H U$. The BW coordinate formula \citep{bhatia_bureswasserstein_2019}
gives
\begin{equation}
\|H\|_A^2
=
\frac12\sum_{i,j}
\frac{|\widetilde H_{ij}|^2}{\lambda_i+\lambda_j}.
\end{equation}
The norm bounds follow from $2m_\delta\le\lambda_i+\lambda_j\le2M_\delta$ and $\|\widetilde H\|_F=\|H\|_F$.
\end{proof}

\paragraph{Interpretation:}
For fixed $d$ and $\delta$, shrinkage keeps the operators uniformly away from singularity and bounds the BW tangent norm relative to the Frobenius norm. These bounds ensure that the inverse and Lyapunov formulas are well defined, but do not guarantee stability of reference eigenspaces when distinct eigenvalues are close. 

\subsection{Finite-sample operator estimation error}
\label{app:operator_estimation_error}

\begin{lemma}[Mean-square error of normalized empirical operators]
\label{lem:operator_mse}
Let $\phi_1,\ldots,\phi_n\in\mathbb R^d$ be independent random vectors satisfying $\|\phi_i\|_2=1$ almost surely. Define
\begin{equation}
\begin{aligned}
Y_i &:= \phi_i\phi_i^\top, &
M_i &:= \mathbb{E}[Y_i], \\
\widehat{\rho}_n &:= \frac{1}{n}\sum_{i=1}^{n}Y_i, &
\bar{\rho}_n &:= \frac{1}{n}\sum_{i=1}^{n}M_i.
\end{aligned}
\end{equation}
Then,
\begin{equation}
\mathbb E\|\widehat\rho_n-\bar\rho_n\|_F^2
=
\frac1{n^2}\sum_{i=1}^n
\left(1-\|M_i\|_F^2\right)
\le\frac1n.
\end{equation}

For fixed $0<\delta<1$, let
\begin{equation}
\begin{aligned}
\widehat{\rho}_{n,\delta}
&= (1-\delta)\widehat{\rho}_n + \frac{\delta}{d}I_d, \\
\bar{\rho}_{n,\delta}
&= (1-\delta)\bar{\rho}_n + \frac{\delta}{d}I_d.
\end{aligned}
\end{equation}
Then,
\begin{equation}
\begin{aligned}
\mathbb{E}\|\widehat{\rho}_{n,\delta}-\bar{\rho}_{n,\delta}\|_F^2
&= (1-\delta)^2\,
   \mathbb{E}\|\widehat{\rho}_n-\bar{\rho}_n\|_F^2 \\
&\le \frac{(1-\delta)^2}{n}.
\end{aligned}
\end{equation}

In the i.i.d.\ case, with $\rho=\mathbb E[\phi\phi^\top]$,
\begin{equation}
\begin{aligned}
\mathbb{E}\|\widehat{\rho}_n-\rho\|_F^2
&= \frac{1-\|\rho\|_F^2}{n} \\
&\le \frac{1-1/d}{n}.
\end{aligned}
\end{equation}
\end{lemma}

\begin{proof}
The proof follows the variance calculation of \citet{muandet_kernel_2016}, applied to the outer-product matrices $\phi_i\phi_i^\top$.

Set $Z_i=Y_i-M_i$. Independence and $\mathbb E[Z_i]=0$ imply $\mathbb E\langle Z_i,Z_j\rangle_F=0$ for $i\ne j$. Moreover, $\|Y_i\|_F^2=\|\phi_i\|_2^4=1$.
Therefore,
\begin{equation}
\begin{aligned}
\mathbb E\|\widehat\rho_n-\bar\rho_n\|_F^2
&=\frac1{n^2}\sum_{i=1}^n\mathbb E\|Z_i\|_F^2\\
&=\frac1{n^2}\sum_{i=1}^n
\left(\mathbb E\|Y_i\|_F^2-\|M_i\|_F^2\right)\\
&=\frac1{n^2}\sum_{i=1}^n
\left(1-\|M_i\|_F^2\right)
\le\frac1n.
\end{aligned}
\end{equation}
The regularized identity follows from
\begin{equation}
\widehat\rho_{n,\delta}-\bar\rho_{n,\delta}
=(1-\delta)(\widehat\rho_n-\bar\rho_n).
\end{equation}
In the i.i.d.\ case, $M_i=\rho$ for every $i$. Substitution gives the stated identity, and $(\operatorname{tr}\rho)^2\le d\|\rho\|_F^2$ gives $\|\rho\|_F^2\ge1/d$.
\end{proof}

\paragraph{Rates and scope:}
Since $\|H\|_{\mathrm{op}}\le\|H\|_F$, the bound also gives
\begin{equation}
\left(
\mathbb E\|\widehat\rho_n-\bar\rho_n\|_{\mathrm{op}}^2
\right)^{1/2}
\le\frac1{\sqrt n},
\end{equation}
with bound $(1-\delta)/\sqrt n$ after applying the same fixed shrinkage to both operators. For observations with different distributions, the target is their average population operator $\bar\rho_n$. The regularized bound compares the empirical and population operators after applying the same shrinkage to both, which means that it does not include the shrinkage bias relative to the unregularized target $\bar\rho_n$. The bound controls operator estimation error but does not by itself establish stability of the component energies, which also depend on the reference eigenspaces.

\subsection{Bures and projection preliminaries}
\label{app:bures_preliminaries}

Throughout this subsection, let $A\in\mathbb S_{++}^d$ and write $A=\sum_{r=1}^m\alpha_rP_r$ for its decomposition into distinct eigenvalues and complete eigenspaces. For $H\in\operatorname{Sym}(d)$, set
\begin{equation}
\begin{aligned}
\Pi_A^{\mathrm{spec}}(H)
&\coloneqq \sum_r P_r H P_r, \\
\Pi_A^{\mathrm{org}}(H)
&\coloneqq H - \Pi_A^{\mathrm{spec}}(H).
\end{aligned}
\end{equation}

Their ranges are the reference-commuting and cross-eigenspace tangent subspaces, respectively:
\begin{equation}
\label{eq:diagnostic-subspaces}
\begin{aligned}
\mathcal{C}_A
&= \{H \in \operatorname{Sym}(d) : [A,H]=0\}, \\
\mathcal{O}_A
&= \{[\Omega,A] : \Omega \in \mathbb{R}^{d\times d},\;
                    \Omega^\top=-\Omega\}.
\end{aligned}
\end{equation}
where \([X,Y]=XY-YX\) denotes the commutator. 

The BW tangent space is \(T_A\mathbb S_{++}^d=\operatorname{Sym}(d)\). The following lemmas collect the standard BW metric, the decomposition into commuting and orbit directions, and the logarithmic displacement, using our notation \citep{dittmann_explicit_1999,bhatia_bureswasserstein_2019}.

\begin{lemma}[Coordinate representation of the Bures metric]
\label{lem:bures-coordinate-metric}
Let $A=U\Lambda U^\top$, where $U$ is orthogonal and $\Lambda=\operatorname{diag}(\lambda_1,\ldots,\lambda_d)$.

For $H,K\in\operatorname{Sym}(d)$, write $\widetilde H=U^\top H U$ and $\widetilde K=U^\top K U$. Then,
\begin{equation}
\label{eq:lem_coor_representation_1}
g_A(H,K)=\frac12\sum_{i,j=1}^d
\frac{\widetilde H_{ij}\widetilde K_{ij}}{\lambda_i+\lambda_j}.
\end{equation}
In particular, \(\|H\|_A^2=\frac12\sum_{i,j}|\widetilde H_{ij}|^2/(\lambda_i+\lambda_j)\).
\end{lemma}

\begin{proof}
In these coordinates, solving \(AX+XA=K\) gives \(\widetilde X_{ij}=\widetilde K_{ij}/(\lambda_i+\lambda_j)\).

Substitution into \(g_A(H,K)=\tfrac12\operatorname{tr}[H\mathcal L_A^{-1}(K)]\) proves the formula. All denominators are positive, so \(g_A\) is a
positive-definite inner product on \(\operatorname{Sym}(d)\). An equivalent form, used in the transport calculation below, is
\begin{equation}
g_A(H,K)=\operatorname{tr}\!\left[
\mathcal L_A^{-1}(H)A\mathcal L_A^{-1}(K)\right].
\end{equation}

Indeed, if \(X=\mathcal L_A^{-1}(H)\) and \(Y=\mathcal L_A^{-1}(K)\), symmetry and cyclicity of trace give \(\tfrac12\operatorname{tr}[(AX+XA)Y]=\operatorname{tr}(XAY)\).
\end{proof}

\begin{lemma}[Commuting and organizational tangent subspaces]
\label{lem:commuting-organizational-subspaces}

The ranges of the two projections are
\begin{equation}
\begin{aligned}
\mathcal C_A
&=\{H\in\operatorname{Sym}(d):[A,H]=0\}\\
&=\operatorname{range}(\Pi_A^{\mathrm{spec}}),\\
\mathcal O_A
&=\{G\in\operatorname{Sym}(d):
P_rGP_r=0\text{ for every }r\}\\
&=\operatorname{range}(\Pi_A^{\mathrm{org}}).
\end{aligned}
\end{equation}

A symmetric matrix $G$ belongs to $\mathcal O_A$ if and only if it can be written as
\begin{equation}
G=[\Omega,A]=\Omega A-A\Omega
\end{equation}
for some skew-symmetric $\Omega\in\mathbb R^{d\times d}$.

Thus $\mathcal O_A$ is the tangent space at $A$ to its orthogonal orbit. Moreover,
\begin{equation}
\operatorname{Sym}(d)=\mathcal C_A\oplus\mathcal O_A.
\end{equation}
\end{lemma}

\begin{proof}
Write $A=U\Lambda U^\top$, where $\Lambda=\operatorname{diag}(\lambda_1,\ldots,\lambda_d)$, and denote matrix coordinates in this eigenbasis by
$\widetilde X=U^\top XU$. Since
\begin{equation}
[\Lambda,\widetilde H]_{ij}
=(\lambda_i-\lambda_j)\widetilde H_{ij},
\end{equation}
we have
\begin{equation}
\label{eq:commutant-coordinate-characterization}
[A,H]=0
\quad\Longleftrightarrow\quad
\widetilde H_{ij}=0
\text{ whenever }\lambda_i\ne\lambda_j.
\end{equation}
These are exactly the matrices retained by $\Pi_A^{\mathrm{spec}}$. The complementary projection $\Pi_A^{\mathrm{org}}$ retains exactly the entries with $\lambda_i\ne\lambda_j$, which characterize matrices with zero within-eigenspace blocks. This proves the two range identities. Every symmetric matrix splits uniquely into these within-eigenspace and cross-eigenspace parts, giving $\operatorname{Sym}(d)=\mathcal C_A\oplus\mathcal O_A$.

We next establish the commutator characterization. For any skew-symmetric $\Omega$, the matrix $[\Omega,A]$ is symmetric and satisfies
\begin{equation}
\bigl(U^\top[\Omega,A]U\bigr)_{ij}
=(\lambda_j-\lambda_i)\widetilde\Omega_{ij}.
\end{equation}
Its entries therefore vanish whenever $\lambda_i=\lambda_j$, so $[\Omega,A]\in\mathcal O_A$.

Conversely, let $G\in\mathcal O_A$ and define
\begin{equation}
\widetilde\Omega_{ij}
=
\begin{cases}
\dfrac{\widetilde G_{ij}}{\lambda_j-\lambda_i},
& \lambda_i\ne\lambda_j,\\[6pt]
0,
& \lambda_i=\lambda_j,
\end{cases}
\qquad
\Omega=U\widetilde\Omega U^\top.
\end{equation}
Since $\widetilde G$ is symmetric, $\widetilde\Omega_{ji}=-\widetilde\Omega_{ij}$, and hence $\Omega^\top=-\Omega$. For distinct eigenvalues, the construction gives
\begin{equation}
(\lambda_j-\lambda_i)\widetilde\Omega_{ij}
=\widetilde G_{ij}.
\end{equation}
For equal eigenvalues, both sides vanish since $G$ has zero within-eigenspace blocks. Thus $[\Omega,A]=G$.

Finally, tangent directions to the orthogonal orbit of $A$ are generated by the curves $e^{t\Omega}Ae^{-t\Omega}$, with $\Omega^\top=-\Omega$. Their initial velocities are
\begin{equation}
\left.\frac{d}{dt}
\bigl(e^{t\Omega}Ae^{-t\Omega}\bigr)\right|_{t=0}
=[\Omega,A],
\end{equation}
identifying $\mathcal O_A$ with the tangent space to this orbit at $A$.
\end{proof}

\begin{lemma}[Bures orthogonality]
\label{lem:bures-orthogonality}
The splitting is orthogonal in the BW inner product:
\begin{equation}
\operatorname{Sym}(d)=\mathcal C_A\oplus_{g_A}\mathcal O_A.
\end{equation}
Thus \(\Pi_A^{\mathrm{spec}}\) and \(\Pi_A^{\mathrm{org}}\) are BW-orthogonal projections onto their respective subspaces.
\end{lemma}

\begin{proof}
For \(H\in\mathcal C_A\) and \(G\in\mathcal O_A\), \(\widetilde H_{ij}=0\) when \(\lambda_i\ne\lambda_j\), whereas \(\widetilde G_{ij}=0\) when \(\lambda_i=\lambda_j\). Thus $\widetilde H_{ij}\widetilde G_{ij}=0$ for every $i,j$, which means that \cref{eq:lem_coor_representation_1}  gives $g_A(H,G)=0$. The direct sum from \cref{lem:commuting-organizational-subspaces} completes the proof.
\end{proof}

\begin{lemma}[Bures transport, logarithmic displacement, and norm identity]
\label{lem:bures-log-norm}

For \(A,B\in\mathbb S_{++}^d\), let \(T=A^{-1/2}(A^{1/2}BA^{1/2})^{1/2}A^{-1/2}\) and \(S=T-I\). Then,
\begin{align}
TAT&=B,\label{eq:finite-pair-transport-identity}\\
\operatorname{Log}^{\mathrm{BW}}_A(B)
&=\mathcal L_A(S)=AS+SA,\label{eq:finite-pair-log-formula}\\
\|\operatorname{Log}^{\mathrm{BW}}_A(B)\|_A^2
&=d_{\mathrm{BW}}^2(A,B).\label{eq:finite-pair-log-distance}
\end{align}
\end{lemma}

\begin{proof}
With \(C=(A^{1/2}BA^{1/2})^{1/2}\), direct multiplication yields \(TAT=A^{-1/2}C^2A^{-1/2}=B\). The minimizing BW geodesic is \citep{bhatia_bureswasserstein_2019}
\begin{align}
\gamma(t)&=((1-t)I+tT)A((1-t)I+tT)\\
        &=(I+tS)A(I+tS).
\end{align}

Its initial velocity is \(\dot\gamma(0)=AS+SA\), establishing the logarithm formula. By the equivalent metric form above,
\begin{equation}
\begin{aligned}
\|\operatorname{Log}^{\mathrm{BW}}_A(B)\|_A^2
&=\operatorname{tr}(SAS)\\
&=\operatorname{tr}(TAT)-2\operatorname{tr}(TA)+\operatorname{tr}(A)\\
&=\operatorname{tr}(B)-2\operatorname{tr}(C)+\operatorname{tr}(A)\\
&=d_{\mathrm{BW}}^2(A,B).
\end{aligned}
\end{equation}
Here \(\operatorname{tr}(TA)=\operatorname{tr}(C)\) follows from cyclicity of trace.
\end{proof}

\subsection{Exact energy allocation and best approximation}
\label{app:proof_prop1}

\begin{proof}[Proof of \cref{thm:finite-pair-decomposition}]
Fix $A,B\in\mathbb S_{++}^d$ and set $V=\operatorname{Log}^{\mathrm{BW}}_A(B)$. Equipped with the BW inner product $g_A$, the finite-dimensional space $\operatorname{Sym}(d)$ is a real Hilbert space \citep{bhatia_bureswasserstein_2019}.

By \cref{lem:commuting-organizational-subspaces,lem:bures-orthogonality}, the projections give
\begin{equation}
\begin{gathered}
V = V_{\mathrm{spec}} + V_{\mathrm{org}}, \\
V_{\mathrm{spec}} \in \mathcal C_A, \qquad
V_{\mathrm{org}} \in \mathcal O_A, \\
\mathcal C_A \perp_{g_A} \mathcal O_A.
\end{gathered}
\end{equation}
Therefore, orthogonality and \cref{lem:bures-log-norm} yield
\begin{equation}
\begin{aligned}
d_{\mathrm{BW}}^2(A,B)
&=\|V\|_A^2\\
&=\|V_{\mathrm{spec}}\|_A^2+\|V_{\mathrm{org}}\|_A^2\\
&=E_{\mathrm{spec}\mid A}(B)+E_{\mathrm{org}\mid A}(B).
\end{aligned}
\end{equation}
Both energies are nonnegative since they are squared norms.

\paragraph{Projection-theorem argument:}
The finite-dimensional linear subspaces $\mathcal C_A$ and $\mathcal O_A$ are closed. The orthogonal projection theorem \citep{kreyszig_introductory_1989}
therefore identifies $V_{\mathrm{spec}}$ and $V_{\mathrm{org}}$ as the unique best approximations of $V$ in these subspaces, respectively.

\paragraph{Direct verification and minimum values:}
For any $H\in\mathcal C_A$, $V_{\mathrm{spec}}-H\in\mathcal C_A$, so orthogonality gives
\begin{equation}
\begin{aligned}
\|V-H\|_A^2
&=\|V_{\mathrm{spec}}-H\|_A^2+\|V_{\mathrm{org}}\|_A^2\\
&=\|V_{\mathrm{spec}}-H\|_A^2+E_{\mathrm{org}\mid A}(B)\\
&\ge E_{\mathrm{org}\mid A}(B).
\end{aligned}
\end{equation}
Since $g_A$ is positive definite, equality holds exactly when $H=V_{\mathrm{spec}}$. Similarly, for any $G\in\mathcal O_A$,
\begin{align}
\|V-G\|_A^2
&=
E_{\mathrm{spec}\mid A}(B)
+\|V_{\mathrm{org}}-G\|_A^2 \\
&\ge E_{\mathrm{spec}\mid A}(B),
\end{align}
with equality exactly when $G=V_{\mathrm{org}}$.

Consequently,
\begin{equation}
\begin{aligned}
E_{\mathrm{org}\mid A}(B)
&=\min_{H\in\mathcal C_A}\|V-H\|_A^2,\\
E_{\mathrm{spec}\mid A}(B)
&=\min_{G\in\mathcal O_A}\|V-G\|_A^2,
\end{aligned}
\end{equation}
with unique minimizers $V_{\mathrm{spec}}$ and $V_{\mathrm{org}}$, respectively.
\end{proof}

\subsection{Spectral energy for trace-one operators}
\label{app:trace_one_spectral_energy}

\begin{proof}[Proof of \cref{thm:spectral-trace-one}]
Let $D=d_{\mathrm{BW}}^2(A,B)$ and $S=T_{A\to B}-I$. We first compute the trace of the logarithmic displacement $V$. Expanding the transport identity gives

\begin{align}
        B&=(I+S)A(I+S)\\
        &=A+V+SAS,\\
        V&=AS+SA.
\end{align}

Since $A$ and $B$ both have trace one, $B-A$ has trace zero. Taking the trace of $B-A=V+SAS$ and using $D=\operatorname{tr}(SAS)$ therefore gives $\operatorname{tr}(V)=-D$.

Now write $V$ in an eigenbasis of $A$. The spectral projection keeps the blocks within each reference eigenspace and removes the blocks between distinct eigenspaces. In particular, it keeps every diagonal entry, so $\operatorname{tr}(V_{\mathrm{spec}})=-D$ as well. To separate this trace from the rest of the spectral displacement, write $V_{\mathrm{spec}}=cA+H$, choosing $c$ so that $\operatorname{tr}(H)=0$. Taking the trace of both sides gives $-D=c$, since $\operatorname{tr}(A)=1$. Thus
\begin{align}
    V_{\mathrm{spec}}&=-DA+H,\\
    H&=V_{\mathrm{spec}}+DA.
\end{align}
Both $V_{\mathrm{spec}}$ and $A$ commute with $A$, so $H$ does too.

It remains to compute the energy of this sum. Since $\mathcal L_A(I/2)=A$, we have $\mathcal L_A^{-1}(A)=I/2$. The BW inner product therefore satisfies
\begin{align}
    g_A(K,A)&=\tfrac12\operatorname{tr}\!\bigl(K\mathcal L_A^{-1}(A)\bigr)\\
    &=\tfrac14\operatorname{tr}(K).
\end{align}

Taking $K=H$ gives $g_A(H,A)=0$, while taking $K=A$ gives $\|A\|_A^2=1/4$. The two terms in $V_{\mathrm{spec}}=-DA+H$ are therefore orthogonal under the BW metric, and their squared norms add:
\begin{align}
    E_{\mathrm{spec}\mid A}(B)
    &= \|-DA+H\|_A^2\\
    &= \frac{D^2}{4}+\|H\|_A^2.
\end{align}
For $A\ne B$, dividing by $D>0$ gives the spectral-share lower bound $D/4$. Equality holds exactly when $H=0$, or equivalently when $V_{\mathrm{spec}}=-DA$.
\end{proof}

\subsection{Commuting zero condition and orthogonal coordinate invariance}
\label{app:proof_prop2}

\begin{proof}[Proof of \cref{prop:organizational-faithfulness}]
Fix \(A,B\in\mathbb S_{++}^d\).

\paragraph{Commuting zero condition:}
Write \(V=\operatorname{Log}^{\mathrm{BW}}_A(B)\), \(T=T_{A\to B}\), and \(S=T-I\). Positive definiteness of the metric and \cref{lem:commuting-organizational-subspaces} imply
\begin{equation}
\begin{aligned}
E_{\mathrm{org}\mid A}(B)=0
&\iff \Pi_A^{\mathrm{org}}(V)=0 \\
&\iff [A,V]=0.
\end{aligned}
\end{equation}

In an eigenbasis of \(A\), the identity \(V=AS+SA\) gives
\begin{equation}
\widetilde V_{ij}
=
(\lambda_i+\lambda_j)\widetilde S_{ij}.
\end{equation}
Since \(\lambda_i+\lambda_j>0\),
\begin{equation}
\begin{aligned}
[A,V]=0
&\iff [A,S]=0 \\
&\iff [A,T]=0.
\end{aligned}
\end{equation}
If \([A,T]=0\), then \(B=TAT=AT^2\) commutes with \(A\). Conversely, commuting real symmetric matrices \(A,B\) are simultaneously orthogonally diagonalizable. In their common eigenbasis,
\begin{equation}
T=\operatorname{diag}\!\left(
\sqrt{\mu_1/\lambda_1},\ldots,\sqrt{\mu_d/\lambda_d}
\right),
\end{equation}
where \(\mu_i\) are the eigenvalues of \(B\) in that basis. Thus \([A,T]=0\), completing the equivalence
\begin{equation}
E_{\mathrm{org}\mid A}(B)=0
\quad\Longleftrightarrow\quad
[A,B]=0.
\end{equation}

\paragraph{Orthogonal coordinate invariance:}
Let \(U\) be orthogonal, \(A'=UAU^\top\), and \(B'=UBU^\top\). Orthogonal equivariance of positive-definite matrix square roots and inverses \citep{horn_matrix_1985}, together with trace invariance, gives
\begin{equation}
\label{eq:orth-inv-distance}
d_{\mathrm{BW}}(A',B')=d_{\mathrm{BW}}(A,B).
\end{equation}
It also gives \(T_{A'\to B'}=UT_{A\to B}U^\top\), and therefore
\begin{equation}
\label{eq:orth-inv-log}
\operatorname{Log}^{\mathrm{BW}}_{A'}(B')
=
U\operatorname{Log}^{\mathrm{BW}}_A(B)U^\top.
\end{equation}

The complete eigenspace projectors of \(A'\) are \(P'_r=UP_rU^\top\). For \(H'=UHU^\top\),
\begin{equation}
\begin{aligned}
\Pi_{A'}^{\mathrm{spec}}(H')
&= \sum_r P'_r H' P'_r \\
&= U \Pi_A^{\mathrm{spec}}(H) U^\top.
\end{aligned}
\end{equation}

Since \(\Pi_A^{\mathrm{org}}=I-\Pi_A^{\mathrm{spec}}\), the organizational projection obeys the same equivariance. Hence
\begin{equation}
V'_k=UV_{\mathrm{k}}U^\top, \quad k\in\{\mathrm{spec,org}\}
\label{eq:orth-inv-vector}
\end{equation}

Finally,
\begin{equation}
\mathcal L_{A'}(UXU^\top)
=
U\mathcal L_A(X)U^\top,
\end{equation}
so
\begin{equation}
\mathcal L_{A'}^{-1}(UHU^\top)
=
U\mathcal L_A^{-1}(H)U^\top.
\end{equation}
Substitution into the BW inner product yields
\begin{equation}
g_{A'}(UHU^\top,UKU^\top)=g_A(H,K).
\end{equation}
Taking squared norms of the transformed components proves
\begin{equation}
\label{eq:orth-inv}
E_{k\mid A'}(B')=E_{\mathrm{k}\mid A}(B), \quad k\in\{\mathrm{spec,org}\}
\end{equation}
When \(A\ne B\), division by the invariant positive total energy also gives
\begin{equation}
\label{eq:orth-inv-org-fraction}
\eta_{\mathrm{org}\mid A'}(B')
=
\eta_{\mathrm{org}\mid A}(B).
\end{equation}
\end{proof}

\begin{remark}[Reference eigenspaces and degeneracy]
The condition \([A,B]=0\) means that \(B\) maps each complete eigenspace of \(A\) into itself. It permits arbitrary symmetric structure within a repeated eigenspace. In particular, if \(A=cI\), then \(\mathcal C_A=\operatorname{Sym}(d)\), \(\mathcal O_A=\{0\}\), and all displacement is spectral. Rotating the basis within a repeated eigenspace leaves its projector unchanged, so the allocation does not depend on that basis choice. Nevertheless, this invariance does not imply stability under perturbations that split or merge eigenvalues.
\end{remark}

\subsection{Recovery of the local Bures decomposition}
\label{app:proof_prop3}

\begin{lemma}[Infinitesimal recovery of the classical Bures decomposition]
\label{prop:local-limit-recovery}
Fix $A\in\mathbb S_{++}^d$ and let $B_\varepsilon\in\mathbb S_{++}^d$ satisfy $B_\varepsilon=A+\varepsilon H+o(\varepsilon)$, with $H\in\operatorname{Sym}(d)$. 

Then
\begin{equation}
\operatorname{Log}^{\mathrm{BW}}_A(B_\varepsilon)
=\varepsilon H+o(\varepsilon),
\end{equation}
and, for \(k\in\{\mathrm{spec},\mathrm{org}\}\),
\begin{equation}
\begin{aligned}
V_{k,\varepsilon}
&= \varepsilon \Pi_A^k(H) + o(\varepsilon), \\
\frac{E_{k\mid A}(B_\varepsilon)}{\varepsilon^2}
&\xrightarrow{\varepsilon\to0} \|\Pi_A^k(H)\|_A^2.
\end{aligned}
\end{equation}
Thus, dividing the component energies by $\varepsilon^2$ recovers the classical local commuting--orbit energies as $\varepsilon\to0$.
\end{lemma}

\begin{proof}
Let \(T_\varepsilon=T_{A\to B_\varepsilon}\) and \(S_\varepsilon=T_\varepsilon-I\). Smoothness of matrix square roots and inverses on \(\mathbb S_{++}^d\) gives \(S_\varepsilon=O(\varepsilon)\).

The transport identity expands as
\begin{equation}
\begin{aligned}
B_\varepsilon
&= (I+S_\varepsilon)A(I+S_\varepsilon) \\
&= A + V_\varepsilon + S_\varepsilon A S_\varepsilon,
\end{aligned}
\end{equation}
where \(V_\varepsilon=\mathcal L_A(S_\varepsilon)\). Since \(S_\varepsilon AS_\varepsilon=O(\varepsilon^2)\),
\begin{equation}
\begin{aligned}
V_\varepsilon
&= B_\varepsilon - A + O(\varepsilon^2) \\
&= \varepsilon H + o(\varepsilon).
\end{aligned}
\end{equation}

Applying the fixed linear projections gives the asserted expansions of the components. Squaring their BW norms gives \(E_{k\mid A}(B_\varepsilon)
=\varepsilon^2\|\Pi_A^k(H)\|_A^2+o(\varepsilon^2)\), which proves the limit.
\end{proof}

\begin{remark}[Interpretation of the local limit]
As $B_\varepsilon$ approaches $A$, the exact logarithmic displacement divided by $\varepsilon$ converges to the tangent direction $H$. Consequently, the component energies divided by $\varepsilon^2$ converge to the corresponding local BW metric contributions on the commuting and orbit tangent subspaces. For references with distinct eigenvalues, these coincide with the decomposition described by
Dittmann \cite{dittmann_explicit_1999}.
\end{remark}

\begin{remark}[Finite-pair construction and local-limit consistency]
\label{rem:finite-pair-as-nonlinear-extension}
The exact allocation in \cref{thm:finite-pair-decomposition} follows from projecting the BW logarithmic displacement onto the classical commuting and orbit subspaces. \cref{prop:local-limit-recovery} shows that this finite-pair construction agrees with the corresponding tangent-space energies in the local limit.
\end{remark}

\section{Visualization conventions}
\label{app:visualization}

We use a consistent color palette adapted from Simplified Science Publishing \citep{simplified_science_publishing_best_nodate}.

\end{document}